\documentclass[sigconf]{acmart}
\AtBeginDocument{%
  }

\copyrightyear{2026}
\acmYear{2026}
\setcopyright{cc}
\setcctype{by}
\acmConference[MM '26] {Proceedings of the 35th ACM International Conference on Multimedia}{November 10--14, 2026}{Rio de Janeiro, Brazil.}
\acmBooktitle{Proceedings of the 35th ACM International Conference on Multimedia (MM '26), November 10--14, 2026, Rio de Janeiro, Brazil}
\acmISBN{979-8-4007-2213-4/2026/11}
\acmDOI{10.1145/3767308.3836288}

\usepackage{multirow}
\usepackage{booktabs}
\usepackage{graphicx}
\usepackage{colortbl}

\usepackage{url}

\usepackage{algorithm}
\usepackage{algorithmic}

\usepackage{subcaption}
\usepackage[table,xcdraw]{xcolor}
\usepackage{tabularx} 
\usepackage{multirow}

\usepackage{placeins}
\usepackage{makecell}
\usepackage{wrapfig}
\usepackage{enumitem}

\usepackage{pifont}

\usepackage{siunitx} 
\usepackage[table]{xcolor} 

\usepackage[most]{tcolorbox}

\usepackage{hyperref}     
\usepackage{fontawesome5} 

\usepackage{bm}       

\newcommand{\R}{\mathbb{R}}

\newcommand{\bh}{\mathbf{h}}

\newcommand{\bS}{\mathbf{S}}

\newcommand{\etal}{\textit{et al.}}
\newcommand{\eg}{\textit{e.g.,}} 
\newcommand{\ie}{\textit{i.e.,}}

\definecolor{color_ours}{RGB}{210, 225, 215}

\definecolor{darkgreen}{RGB}{0,100,0}

\newcolumntype{Y}{>{\centering\arraybackslash}X}

\usepackage[capitalize,noabbrev]{cleveref}

\theoremstyle{plain}

\newtheorem{definition}{Definition}

\newtheorem{corollary}{Corollary}

\begin{document}

\title{Disentangling Semantic Attention from Structural Bias in the Attention Manifold}

\author{Pengkun Jiao}
\affiliation{%
  \institution{Fudan University}
  \city{Shanghai}
  \country{China}
}
\email{pkjiao23@m.fudan.edu.cn}

\author{Bin Zhu}
\affiliation{%
  \institution{Singapore Management University}
  \country{Singapore}
}
\email{binzhu@smu.edu.sg}

\author{Jingjing Chen}
\authornote{Corresponding author.}
\affiliation{%
  \institution{Fudan University}
  \city{Shanghai}
  \country{China}
}
\email{chenjingjing@fudan.edu.cn}

\author{Yu-Gang Jiang}
\affiliation{%
  \institution{Fudan University}
  \city{Shanghai}
  \country{China}
}
\email{ygj@fudan.edu.cn}

\renewcommand{\shortauthors}{P Jiao et al.}



\begin{abstract}
The empirical success of attention mechanism in Multimodal Large Language Models (MLLMs) often obscures its inherent, subtle flaws. 
Specifically, MLLMs consistently exhibit disproportionate attention toward certain semantically uninformative visual tokens, a phenomenon termed “register” or “Visual Attention Sinks.” While existing inference intervention methods attempt to identify these sink tokens and redistribute their attention weights, such approaches typically treat these tokens in isolation and suffer from computational inefficiency. Instead, we reframe this phenomenon as a generalized textual bias exerted over visual features that extends beyond isolated sink tokens. From this perspective, a pervasive structural bias leads to the dilution of the semantic visual signal, precipitating multimodal hallucinations as the model prioritizes linguistic priors over valid visual evidence. To address this limitation, we introduce Saliency-guided Purification and Adaptive Redistribution (SPAR), a training-free, plug-and-play intervention. SPAR mitigates this generalized textual bias by purifying structural noise and subsequently redistributing the reclaimed attention budget to the most informative visual regions. Comprehensive evaluations across a diverse spectrum of hallucination benchmarks demonstrate that SPAR effectively restores authentic visual grounding with negligible computational overhead. \href{https://github.com/pengkun-jiao/SPAR}{\faGithub~Project Page}
\end{abstract}

\ccsdesc[500]{Computing methodologies~Computer vision}
\ccsdesc[500]{Computing methodologies~Natural language processing}

\keywords{multimodal large language models, attention, structural bias, hallucination, posterior collapse, linguistic prior}


\maketitle

\begin{figure}[t!]
    \centering
    \vspace{0.1in}
    \includegraphics[width=0.88\linewidth]{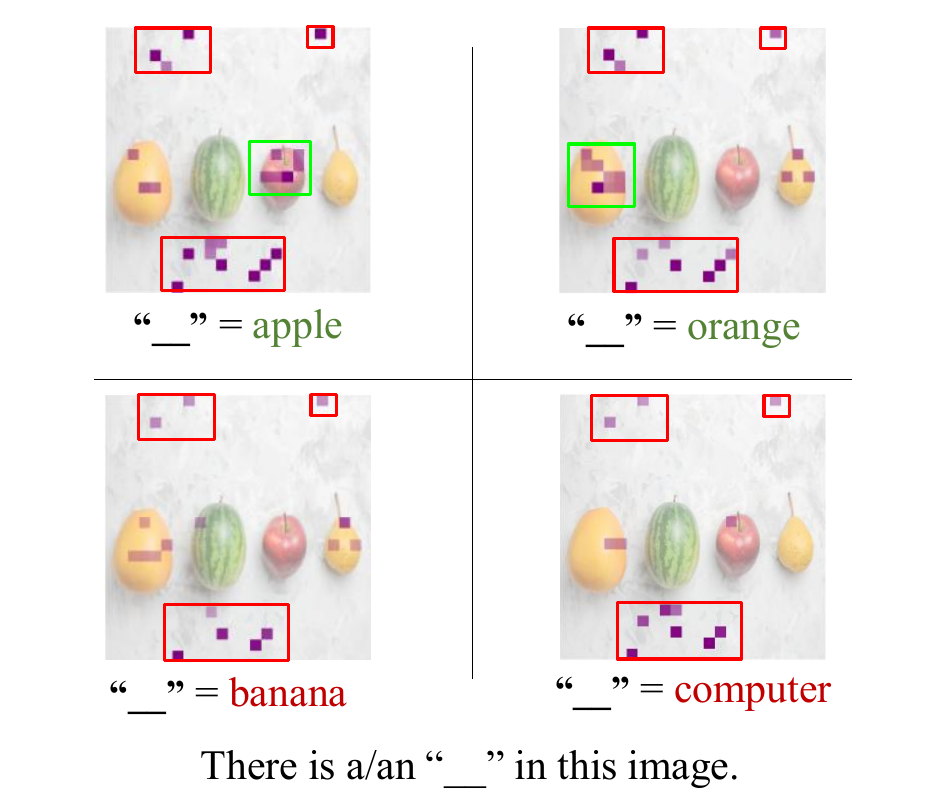}
    \vspace{-0.1in}
    \caption{Visualization of structural attention noise. High attention responses consistently concentrate on fixed image regions across different textual queries, indicating a query-invariant bias.}
    \label{fig:fixed_attention_sink}
\end{figure}

\section{Introduction}

The rapid advancement and successful deployment of Multimodal Large Language Models (MLLMs) across diverse real-world applications have highlighted their transformative potential~\cite{li2024llava_next,wang2024qwen2vl,chen2024internvl,jiao2025fhtc,jiao2026rode}. However, their overall reliability is persistently compromised by hallucinations—where models produce responses inconsistent with the visual evidence—significantly hindering their safe and dependable adoption~\cite{bai2024hallucination,li2023pope}. 

Traditionally, literature has focused on \textit{Spontaneous Hallucinations}~\cite{li2023pope,rohrbach2018chair}, which arise from internal model noise or an over-reliance on statistical priors. 
Yet, as MLLMs are increasingly deployed in complex, interactive environments, a more severe vulnerability has emerged: \textit{Induced Hallucinations}, or \textit{Gaslighting}~\cite{zhu2025gaslightBench}. In these scenarios, misleading external cues, \eg deceptive user prompts, compel the model to abandon its visual grounding entirely, prioritizing linguistic compliance over visual fidelity~\cite{zhu2025gaslightBench}.

An investigation into the internal dynamics of MLLMs reveals a structural flaw: MLLMs frequently exhibit persistent, query-agnostic attention artifacts. As illustrated in Figure~\ref{fig:fixed_attention_sink}, rather than focusing on task-relevant objects, the model's attention is often misdirected toward fixed spatial regions or semantically uninformative tokens. Recent studies have characterized this behavior as \textit{registers}~\cite{darcet2024vit_need_register} or \textit{visual attention sinks}~\cite{kang2025see_what_you_are_told}, treating these anomalous attention patterns as isolated receptacles to absorb redundant attention.
However, we argue that these identified sinks are merely localized symptoms of a broader phenomenon: a generalized text-to-visual attention bias.

\begin{figure}[t]
     \centering
     \begin{subfigure}[b]{0.23\textwidth}
         \centering
         \includegraphics[width=\textwidth]{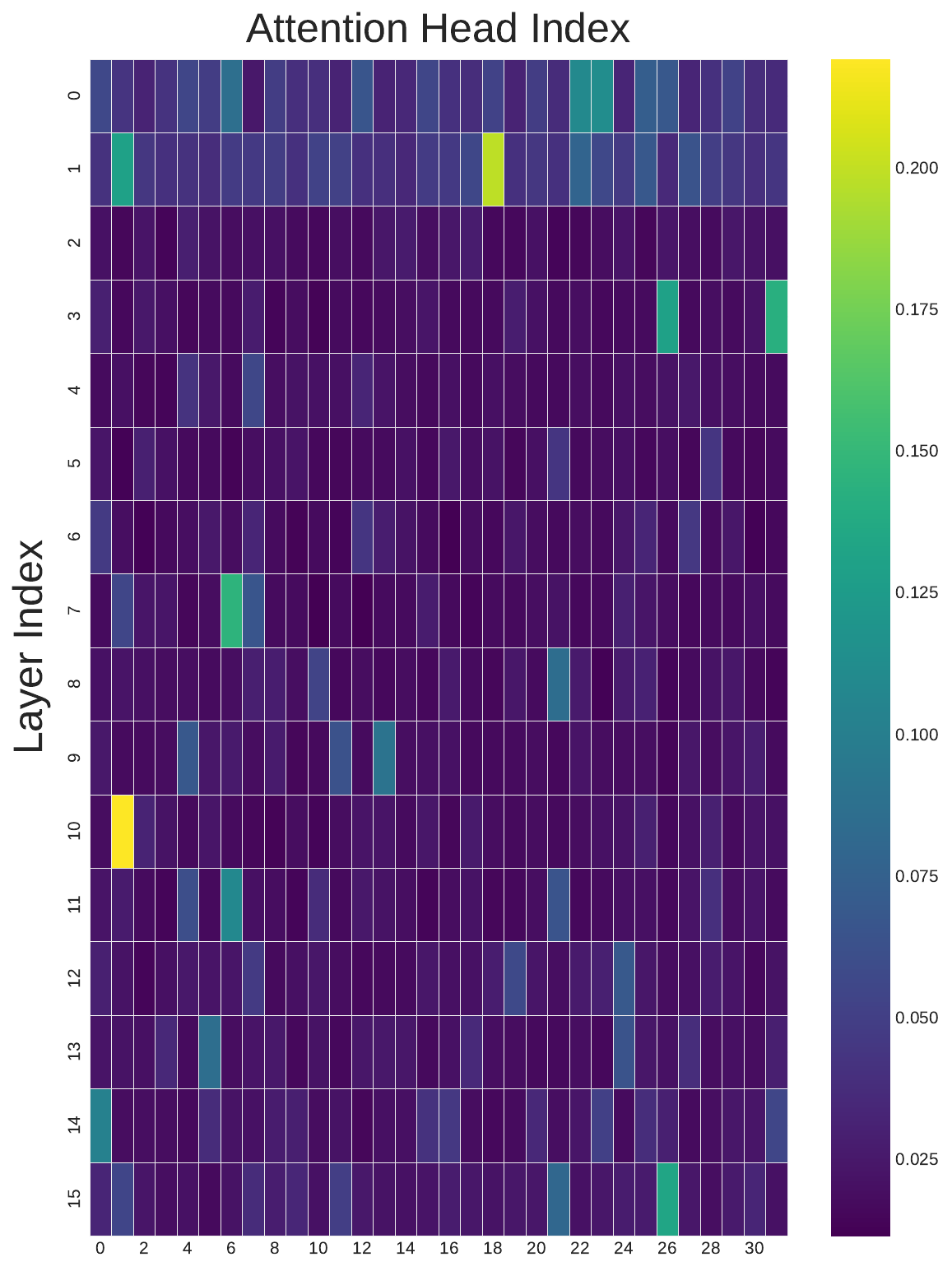}
         \caption{Text-to-Text Attention}
         \label{fig:image1}
     \end{subfigure}
     \hfill 
     \begin{subfigure}[b]{0.23\textwidth}
         \centering
         \includegraphics[width=\textwidth]{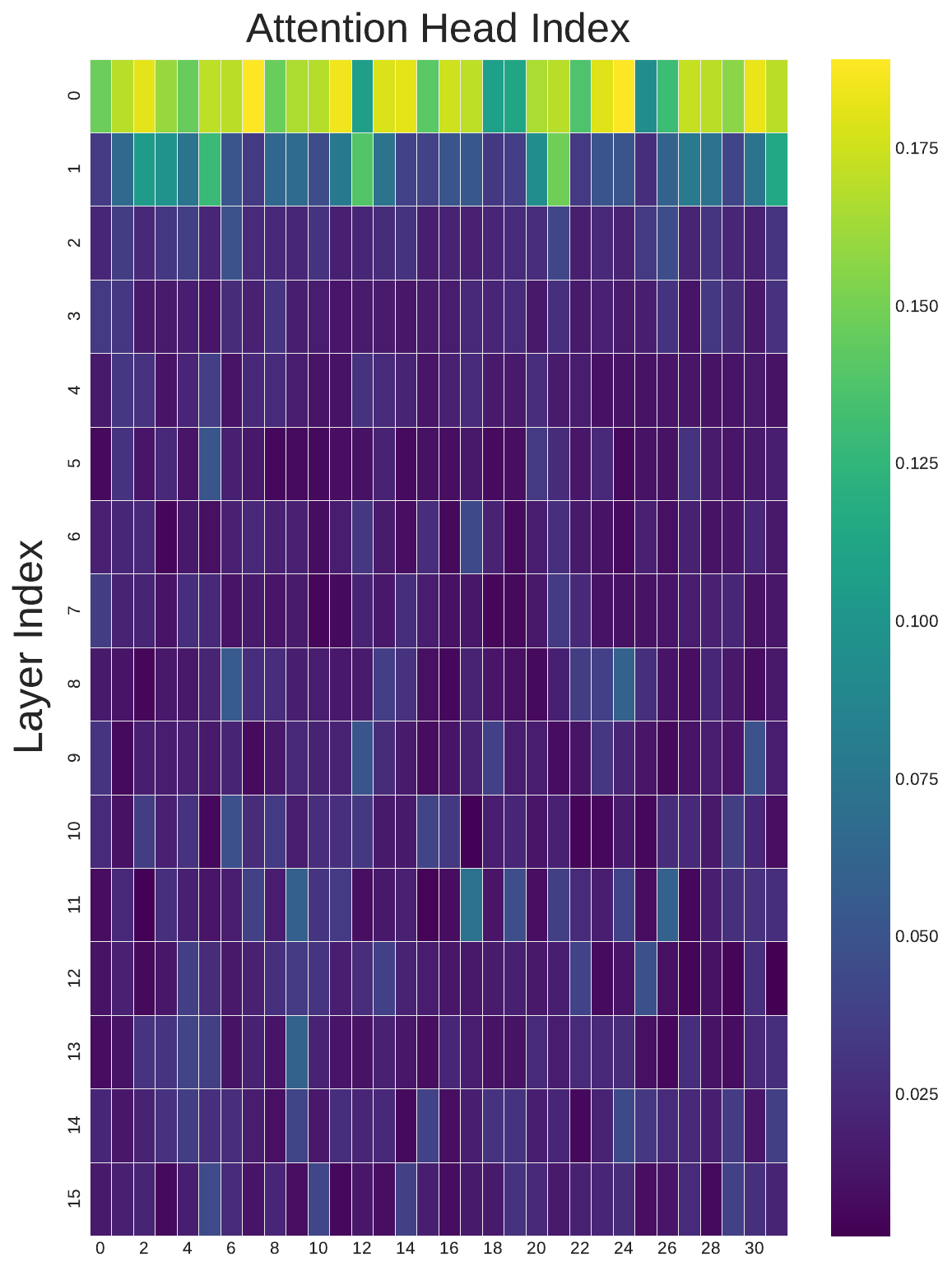}
         \caption{Text-to-Image Attention}
         \label{fig:image2}
     \end{subfigure}
     \vspace{-0.1in}
     \caption{Mean attention maps of (a) text-to-text and (b) text-to-image interactions in LLaVA-NeXT. Structural bias is substantially stronger in text-to-image attention, especially in the first two layers.} 
     \label{fig:token_attention}
\end{figure}

We posit that the attention manifolds in MLLMs exhibit a static, latent \textit{Structural Bias}. Mathematically, this stems from the softmax normalization constraint, which forces the model to distribute residual probability mass across non-informative regions to satisfy the unit-sum requirement. By analogy to Polya’s Urn dynamics~\cite{mahmoud2008polya}, we hypothesize that the model reinforces subtle initial asymmetries—such as those inherent in rotary positional encodings~\cite{su2024rope}—until they crystallize into fixed positional biases. As shown in Figure~\ref{fig:token_attention}, text tokens maintain a high average attention toward fixed image positions across initial layers, where visual and linguistic modalities have not yet deeply interacted, allowing these asymmetric structural priors to easily dominate.
Consequently, the \textit{Semantic Attention} signal is eclipsed by this high-magnitude structural bias, rendering the model highly susceptible to hallucination.

Current inference-time interventions exhibit critical limitations. Contrastive decoding methods~\cite{leng2024vcd,chen2024halc} typically require multiple forward passes per generated token, fundamentally slowing down generation. More recent attention-calibration strategies~\cite{kang2025see_what_you_are_told,jiao2025gasEraser} successfully mitigate hallucinations by reallocating attention away from isolate non-essential “sink” tokens; however, they also introduce substantial inference latency, limiting their practicality for latency-sensitive applications. Furthermore, localized interventions that merely mask or sequester discrete sink tokens fail to address the underlying generalized structural biases.


\begin{figure}[htbp]
    \vspace{0.1in}
    \centering
    \begin{subfigure}{0.47\linewidth}
        \centering
        \includegraphics[width=1.0\linewidth]{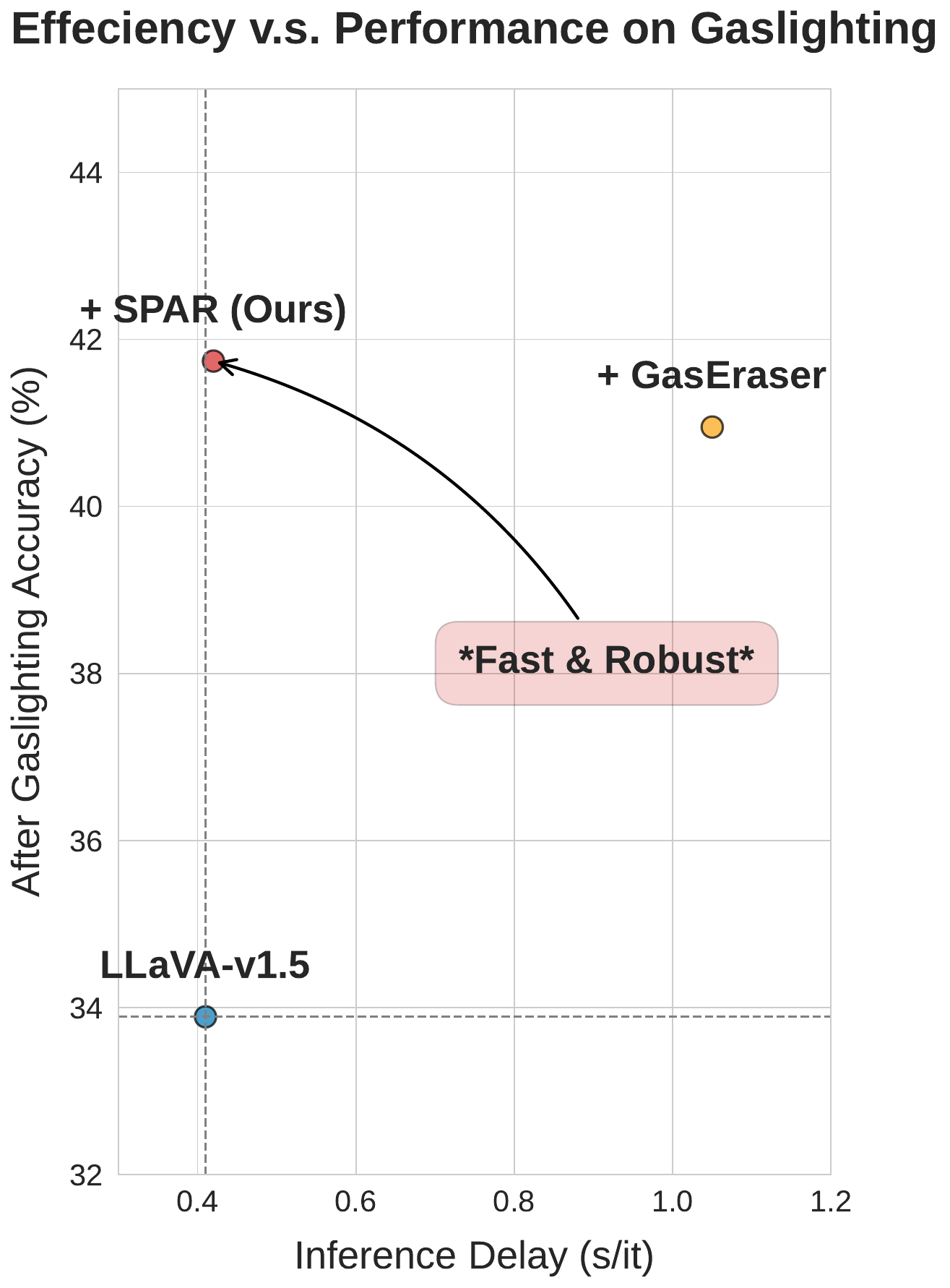}
        \caption{}
    \end{subfigure}\hfill
    \begin{subfigure}{0.46\linewidth}
        \centering
        \includegraphics[width=1.0\linewidth]{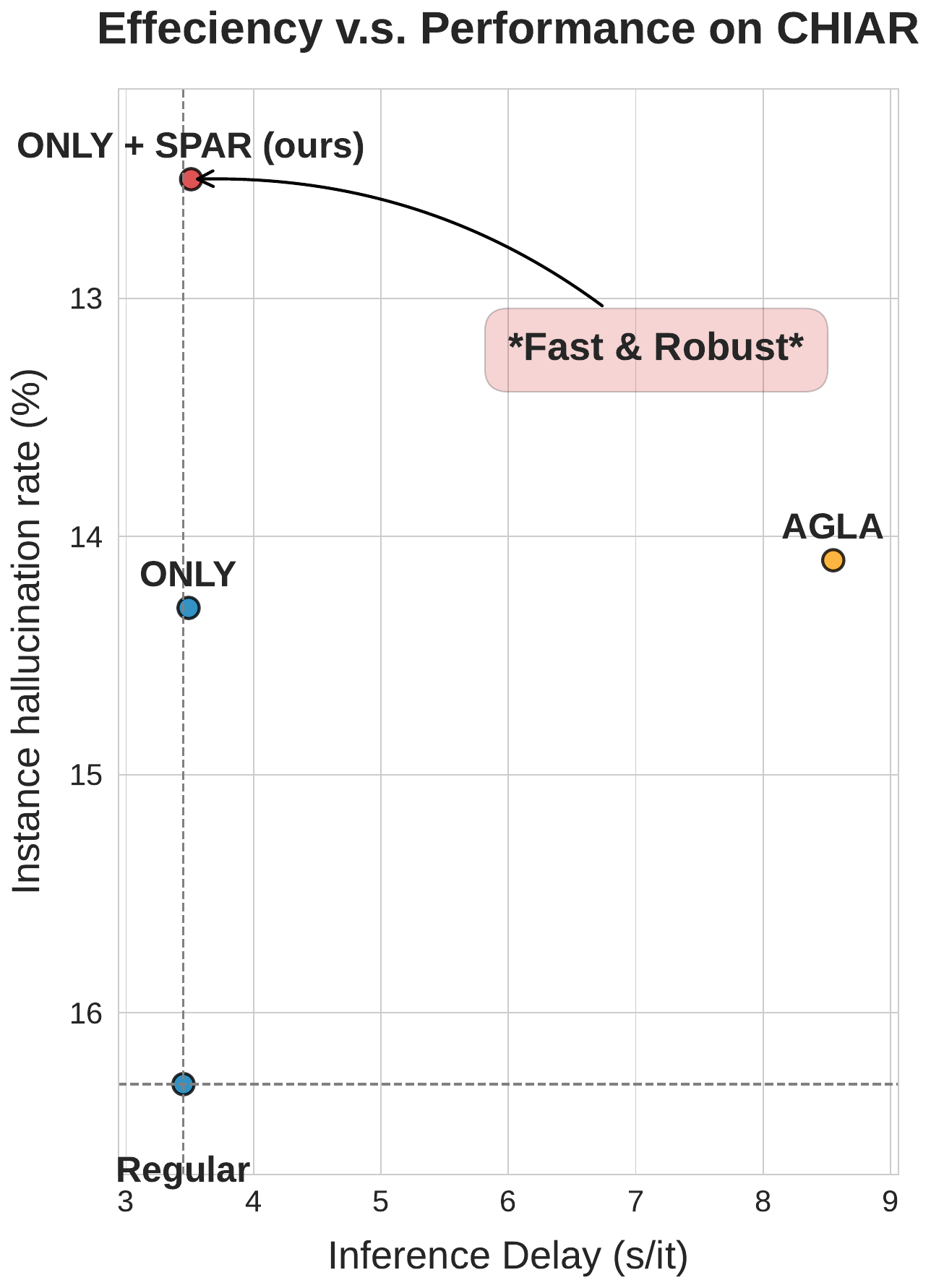}
        \caption{}
    \end{subfigure}
    \vspace{-0.1in}
    \caption{SPAR improves robustness to both (a) spontaneous hallucinations and (b) induced hallucinations while incurring minimal inference overhead.}
    \label{fig:performance}
\end{figure}

To address these issues, we propose a Signal-Noise Decomposition perspective, asserting that robust grounding requires decoupling rigid structural bias from attention manifold. Building on this insight, we introduce \textbf{S}aliency-guided \textbf{P}urification and \textbf{A}daptive \textbf{R}edistribution (\textbf{SPAR}). Unlike prior methods that target isolated sinks, SPAR adopts a holistic approach by treating structural bias as a generalized noise layer superimposed on the attention manifold. 
As shown in Figure~\ref{fig:performance}, evaluations confirm that SPAR provides a highly efficient, training-free solution that substantially enhances MLLM robustness against both spontaneous hallucinations and induced gaslighting, bridging the gap between cross-modal fidelity and low-latency inference demands.

Our main contributions can be summarized as:
\begin{itemize} [nosep]
\item[\ding{182}] We formalize the Semantic-Structural Decoupling hypothesis, demonstrating that generalized structural biases act as static noise that dilutes semantic signals.
\item[\ding{183}] We propose SPAR, a training-free, plug-and-play mechanism that leverages visual saliency to decouple structural bias and restore visual SNR.
\item[\ding{184}] We demonstrate SPAR’s effectiveness across diverse benchmarks, achieving competitive performance in mitigating both spontaneous and induced hallucinations with negligible latency.
\end{itemize}

\section{Related Works}

\paragraph{Attention Sink} describes the tendency of LLMs to allocate disproportionately high attention to a small subset of tokens—typically the initial few—regardless of their semantic contribution~\cite{xiao2024stream_llm}. Xiao et al.~\cite{xiao2024stream_llm} demonstrated that this behavior is consistent and gives rise to a strong attentional bias toward these early tokens.
Subsequent research has sought to uncover the underlying causes of this phenomenon. Cancedda et al.\cite{cancedda-2024-spectral} proposed that the attention sink is often concentrated in the very first token, attributing this bias to the large norm of its hidden state. In contrast, other studies have observed that attention sinks can manifest in various semantically weak tokens without a fixed position in the input sequence~\cite{sun2024massive, yu2024unveiling}, complicating the attention distribution. 
A similar phenomenon, referred to as \textit{Visual Attention Sink}, has also been observed in MLLMs. Timothée \etal~\cite{darcet2024vit_need_register} showed that high-norm tokens frequently emerge during inference in vision transformers, primarily from low-information background regions of images. Seil \etal~\cite{kang2025see_what_you_are_told} further investigated this phenomenon in MLLMs and proposed an attention reallocation strategy to mitigate it.

\paragraph{Hallucination in MLLMs.}
MLLMs may generate responses that hallucinate content not grounded in the visual input, resulting in misinformation and degraded performance~\cite{li2023pope,leng2024vcd}.
Early attempts to address this issue have relied on additional robust instruction tuning using specially curated datasets~\cite{jiang2024hallu_data_1,sun2024aligning}. Although these approaches can be effective, they demand substantial training resources, making them impractical for most users.
Consequently, recent efforts have shifted toward training-free inference-time interventions. One dominant category is contrastive decoding–based methods, which mitigate hallucinations by contrasting logits from paired outputs~\cite{chen2024halc,leng2024vcd,favero2024visual_infor_grounding,an2024agla}. While promising, these methods typically require multiple queries, slowing down generation, though a very recent work~\cite{wan2025only} attempts to optimize this via a single-query, one-layer intervention.
Parallel to logit-level manipulation, another emerging line of research focuses on refining internal attention mechanisms and visual grounding. For example, PAI~\cite{liu2024pai} directly enhances visual fidelity by compelling the model to pay more attention to image tokens during decoding without any extra training. To further interpret these internal dynamics, HGAI~\cite{jiang2025hgai} utilizes an attention lens to identify and rectify object hallucinations originating in the middle layers of MLLMs. From a causal perspective, CasualMM~\cite{zhou2024CasualMM} mitigates hallucinations by deciphering attention causality to suppress modality priors that often lead to erroneous predictions. Moreover, MemVR~\cite{zou2025memvr} introduces a memory-space visual retracing mechanism, enabling the model to “look twice" at the visual input to verify its responses.
More recently, attention has turned toward induced hallucinations triggered by misleading prompts. GaslightBench~\cite{zhu2025gaslightBench} established a benchmark for evaluating model resilience against linguistic pressure (i.e., gaslighting). To address this, GasEraser~\cite{jiao2025gasEraser} was proposed to mitigate these effects by relocating anomalous attention patterns specifically triggered under such pressure.
GasVideo~\cite{tang2025spatiotemporal} further extends negation-based gaslighting to video LLMs.


\section{Semantic-Structural Decoupling}
\label{sec:theory}

In this section, we formalize the mechanism of visual hallucination through the lens of Information Theory and Bayesian Inference. We propose the \textbf{Semantic-Structural Decoupling (SSD)} perspective, which posits that the attention manifold is a superposition of latent semantic attentions and rigid structural bias. We demonstrate that hallucinations become inevitable when the rigid structural bias overshadows the semantic signal.

\subsection{Preliminaries: MLLMs and Softmax Attention}

Multimodal Large Language Models (MLLMs), such as LLaVA~\cite{liu2024llava1.5}, typically integrate a visual encoder $f_{\phi}$, a modality projector $g_{\theta}$, and a Transformer-based LLM $h_{\psi}$.

Given an image $I$, the visual features are mapped into the word embedding space as tokens $X_v = \{x_{v,1}, \dots, x_{v,n}\}$, which are concatenated with textual tokens $X_t$ to form the input sequence $X = [X_v; X_t]$.
These tokens are processed via softmax attention. For a given head, the attention matrix $A = [\alpha_{ij}]$ is computed as:
\begin{align} \label{eq:softmax_attention}
    A = \text{Softmax}\left( \frac{QK^T}{\sqrt{d_k}} \right), \text{where } \alpha_{ij} = \frac{\exp(q_i k_j^T / \sqrt{d_k})}{\sum_{l} \exp(q_i k_l^T / \sqrt{d_k})}.
\end{align}
The softmax operation imposes a unit-sum constraint, \ie $\sum_{j} \alpha_{ij} = 1$, which forces attention mass to be allocated even when the query has weak semantic correspondence with the available visual tokens.
Furthermore, the Exponential Amplification property~\cite{zuhri2025softpick} of Softmax exacerbates this issue by magnifying even minor variations in key norms (e.g., positional encoding asymmetry). Ultimately, this causes high-variance visual tokens to disproportionately dominate the attention distribution, emerging as attention `sinks'~\cite{kang2025see_what_you_are_told}.

\begin{figure*}[htbp]
    \centering
    \includegraphics[width=0.95\linewidth]{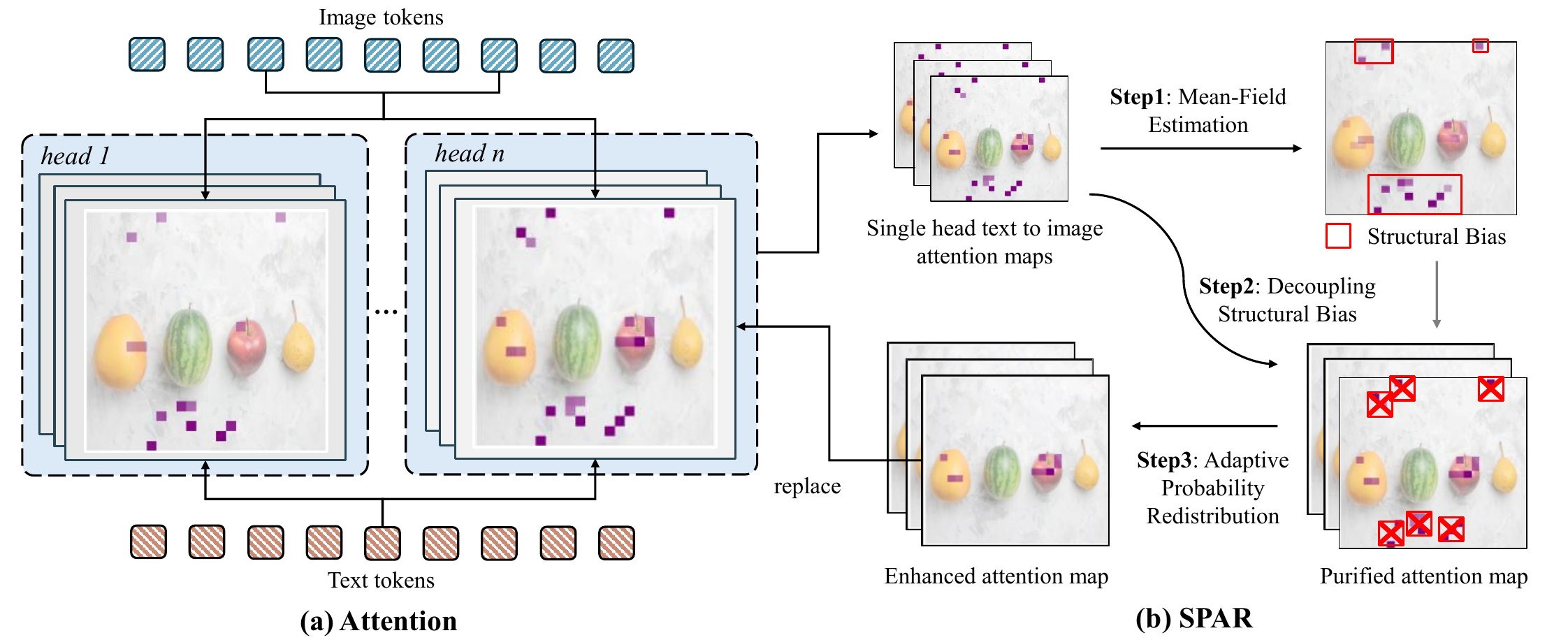}
    \vspace{-0.1in}
    \caption{\textbf{Illustration of SPAR Pipeline.} 
    (a) Text-to-image cross-attention maps are extracted from all attention heads. 
    (b) A fixed structural bias is derived from high-frequency patterns within the attention maps. This bias is then subtracted to purify the attention distribution, and the residual probability mass is adaptively redistributed to the remaining regions to reinforce visual attention.}
\label{fig:SPAR_pipeline}
    \label{fig:framework}
\end{figure*}

\subsection{Manifold Decomposition: The Signal-Noise Hypothesis}
\label{sec:definitions}

The unit-sum constraint in standard Softmax forces the model to distribute probability mass even in the absence of correlation. Consequently, the resulting attention map may conflate true semantic importance with normalization artifacts.
To formalize this, we model the observed attention matrix $\mathbf{A}_{obs} \in \mathbb{R}^{|T| \times |V|}$ as a superposition of a latent semantic signal and a structural bias component.


\begin{definition}[\textbf{Structural Bias}]
We define \textit{structural bias}, denoted by $\mathbf{A}_{struct}$, as the query-invariant component of the attention manifold. It captures the inherent spatial preference of an attention head independent of the specific text query:
\begin{equation}
    \mathbf{A}_{struct}(V) \triangleq \mathbb{E}_{q \sim P(T)} [\mathbf{A}_{obs}(q, V)] \approx \frac{1}{N} \sum_{i=1}^N \mathbf{A}_{obs}(i, \cdot)
\end{equation}
\end{definition}


\begin{definition}[\textbf{Semantic Attention}]
We define \textit{semantic attention}, denoted by $\mathbf{A}_{sem}^*$, as the latent alignment distribution that reflects the true semantic correspondence between textual queries and visual features, disentangled from structural bias.
\end{definition}

Based on the above definitions, we posit an additive noise model for the observed attention:
\begin{equation} \label{eq:attention_decomp}
    \mathbf{A}_{obs}(T, V) = \underbrace{\mathbf{A}_{sem}^*(T, V)}_{\text{True Signal}} + \underbrace{\lambda \cdot \mathbf{A}_{struct}(V)}_{\text{Structural Bias}} + \;\zeta ,
\end{equation}
where $\lambda$ is a modulation coefficient capturing the varying susceptibility to structural bias across different attention heads. The term $\zeta$ represents the \textit{Stochastic Approximation Error}, encompassing approximation errors and high-entropy noise that, while not part of the query-invariant structural bias, contributes no discriminative semantic information to the cross-modal alignment.
Further empirical analysis of structural bias is detailed in Appendix~\ref{appen:further_ana_of_tvab}.

\subsection{Inference as Iterative Bayesian Update}
We interpret the autoregressive generation process of MLLMs through the lens of iterative Bayesian updating. For a candidate next-token $y \in \mathcal{Y}$, the posterior probability is given by:
\begin{equation}
    P(y|V, T) = \frac{P(V|y, T) \cdot P(y|T)}{\sum_{y' \in \mathcal{Y}} P(V|y', T) \cdot P(y'|T)},
    \label{eq:bayes}
\end{equation}
where:
\begin{itemize}
    \item $P(y|T)$ is the \textbf{Linguistic Prior}, derived from the LLM's pre-trained corpus and the current textual context (including system instructions and user prompts).
    \item $P(V|y, T)$ is the \textbf{Visual Likelihood}, representing the probability that the visual evidence $V$ supports the hypothesis of token $y$.
\end{itemize}

\subsection{Posterior Collapse}

To quantify the relative visual support across the vocabulary $\mathcal{Y}$, we define the normalized visual likelihood
\begin{equation}
L(y)=\frac{P(V|y,T)}{\sum_{y'\in\mathcal{Y}}P(V|y',T)}.
\end{equation}

Let $\mathbf{c}_y$ denote the visual context vector aggregated for evaluating candidate token $y$. When the attention manifold is dominated by structural bias $\mathbf{A}_{struct}$ and residual noise $\zeta$, the resulting visual context becomes increasingly insensitive to token-specific semantics. In that regime, the visual likelihood loses discriminability across competing candidates. To characterize this degradation, we define the visual signal-to-noise ratio (SNR) as
\begin{equation}
\mathrm{SNR}\triangleq \frac{\|\mathbf{A}_{sem}^*\|_F^2}{\|\lambda \mathbf{A}_{struct}+\zeta\|_F^2},
\end{equation}
where $\| \cdot \|_F$ denotes the Frobenius norm, representing the total energy concentrated in each component of the attention manifold.



\begin{corollary}[Posterior Collapse Toward the Linguistic Prior]
Assume that, as $\mathrm{SNR}\to 0$, the normalized visual likelihood $L(y)$ becomes approximately non-discriminative across candidate tokens, i.e., there exists a continuous function $\varepsilon(\mathrm{SNR})\ge 0$ with $\lim_{\mathrm{SNR}\to 0}\varepsilon(\mathrm{SNR})=0$ such that
\begin{equation}
    \sup_{y,y'\in\mathcal{Y}} |L(y)-L(y')| \le \varepsilon(\mathrm{SNR}).
\end{equation}
Then the posterior distribution approaches the linguistic prior, and
\begin{equation}
    D_{KL}(P(y|V,T)\parallel P(y|T)) \le f(\mathrm{SNR}),
\end{equation}
where $f(\cdot)$ is non-negative, continuous, and satisfies $\lim_{\mathrm{SNR}\to 0} f(\mathrm{SNR})=0$.
\end{corollary}

The proof sketch of Corollary 1 is provided in Appendix~\ref{appen:proof}. This corollary provides a unified explanation for both spontaneous and induced hallucinations: 
\begin{itemize} [leftmargin=*, nosep]
    \item \textbf{Spontaneous Hallucination:} This phenomenon occurs when the visual likelihood $P(V|y,T)$ is attenuated by internal structural bias. As the $SNR$ decreases, the visual evidence fails to produce a discriminative distribution, causing the posterior to collapse toward the \textit{Linguistic Prior} $P(y|T)$. In this state, the generative process is governed by the statistical density of the pre-training corpus rather than the provided visual features.
    
    \item \textbf{Induced Hallucination:} This occurs when a deceptive textual context $T_{adv}$ biases the linguistic prior $P(y|T_{adv})$ toward a specific candidate subspace. Under structural saturation (low $SNR$), the low-magnitude visual likelihood is insufficient to counteract the high-probability mass of the biased prior. 
\end{itemize}

\begin{figure*}[ht]
\centering
\includegraphics[width=\linewidth]{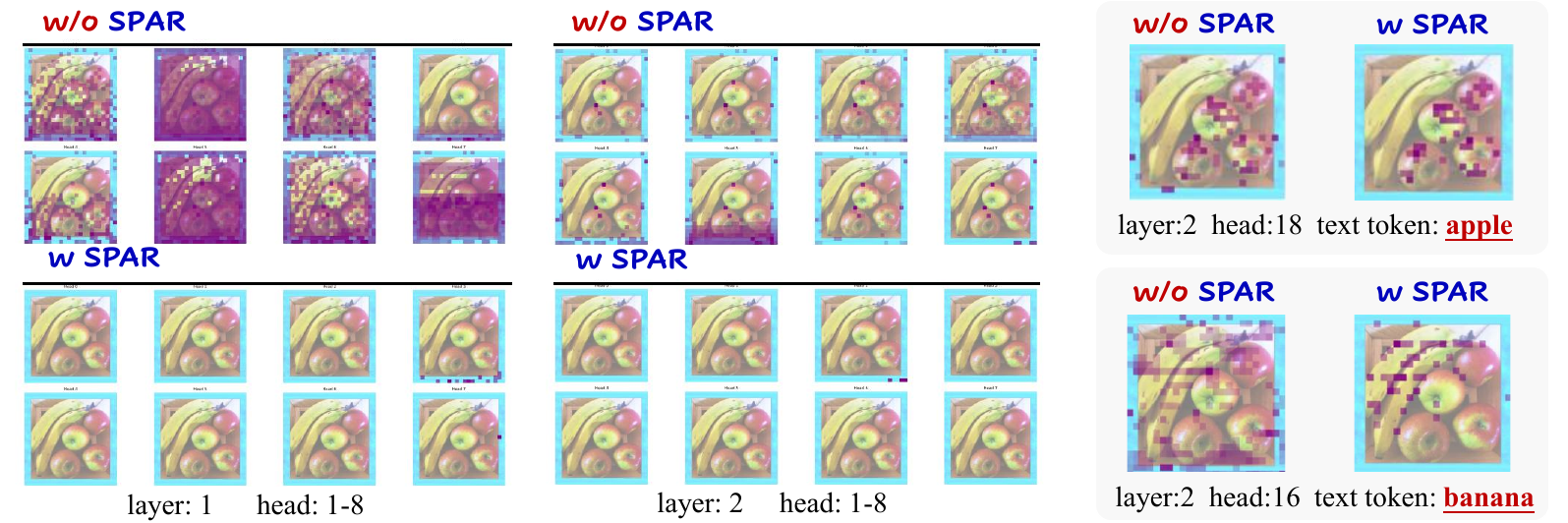}
\vspace{-0.2in}
\caption{{Qualitative visualization of attention purification via SPAR on LLaVA-v1.5.} (Left/Middle) Average text-to-image attention maps reveal that the baseline model suffers from persistent, semantically irrelevant biases, which are effectively rectified by SPAR. (Right) Token-level analysis demonstrates that SPAR suppresses structural bias and sharpens the attention focus on corresponding visual entities.}
\label{fig:text_2_img_atten}
\end{figure*} 

\section{Methodology}

A straightforward strategy for mitigating the degradation of visual discriminability is to isolate and remove the deterministic structural bias inherent in the attention manifold. To this end, we propose \textbf{SPAR} (\textbf{S}aliency-guided \textbf{P}urification and \textbf{A}daptive \textbf{R}edistribution). By purifying the attention manifold, \textbf{SPAR} effectively increases the $SNR$ of the visual likelihood, thereby preventing the posterior distribution from collapsing toward the linguistic prior.

Given the cross-modal attention mechanism in an MLLM layer, let $\mathbf{T}$ and $\mathbf{V}$ denote the sets of text and visual tokens, respectively. The attention matrix $\mathbf{A} \in \mathbb{R}^{|T| \times |V|}$ represents the correlation between text queries and visual keys. We observe that in early layers, $\mathbf{A}$ often exhibits a ``spurious template'' pattern, where text tokens uniformly attend to high-variance visual regions regardless of actual semantic alignment. 

A straightforward intervention would involve the global suppression of high-frequency attention; however, such a naive approach introduces two critical risks. First, it may lead to semantic erasure, inadvertently degrading visual signals that must remain concentrated to support multiple textual tokens (\eg "a red silk skirt"). Second, it risks noise amplification by indiscriminately elevating low-frequency noise attention (\ie $\zeta$). To address these challenges, we propose a two-stage framework: \textbf{Saliency-Guided Structural Bias Purification} (Sec.~\ref{sec:bias_remove}) and \textbf{Adaptive Probability Redistribution} (Sec.~\ref{sec:attention_redis}).

\subsection{Saliency-Guided Structural Bias Purification}
\label{sec:bias_remove}

Our method relies on the hypothesis that visual tokens with extremely high variance in the feature space act as attention sinks\cite{sun2024massive,kang2025see_what_you_are_told}. We quantify this using a saliency map derived from the visual hidden states $\bh \in \R^{|V| \times D}$.

For each visual token $j$, we calculate the feature variance $\sigma^2_j$. To ensure numerical stability and comparability across batches, we normalize this variance:
\begin{equation}
    \sigma^2_j = \text{Var}(\bh_j), \quad 
    \hat{v}_j = \frac{\sigma^2_j - \mu_{\sigma}}{\delta_{\sigma} + \epsilon},
\end{equation}
where $\mu_{\sigma}$ and $\delta_{\sigma}$ are the mean and standard deviation of the variances across the visual sequence, and $\epsilon$ is a small constant (e.g., $1e^{-6}$). The saliency score $\bS_j \in (0, 1)$ is then obtained via a sigmoid activation:
\begin{equation}
    \bS_j = \frac{1}{1 + e^{-\hat{v}_j}}.
\end{equation}

Then we estimate Structural Bias Template $\bar{\mathbf{A}}$ by marginalizing over the textual dimension of $\mathbf{A}$:
\begin{equation}
    \bar{\mathbf{A}}_j = \mathbb{E}_{i \sim T}[\mathbf{A}_{i,j}] \approx \frac{1}{|T|} \sum_{i=1}^{|T|} \mathbf{A}_{i,j}.
\end{equation}
This template captures the persistent spatial preferences of the attention head, where high values in $\bar{\mathbf{A}}_j$ indicate visual regions acting as ``attractors'' regardless of query semantics. To purify the attention $\mathbf{A}$, we subtract this structural bias. However, to avoid harming genuine visual perception (e.g., background context), we modulate the subtraction using the visual feature saliency $\mathbf{S}$:
\begin{equation}
    \mathbf{A}^{pur}_{i,j} = \text{ReLU} \left( \mathbf{A}_{i,j} - \lambda \cdot (1 + \beta \mathbf{S}_j) \cdot \bar{\mathbf{A}}_j \right),
\end{equation}
where $\lambda$ and $\beta$ are hyperparameters. This operation effectively decouples the structural bias from the attention map, exposing the underlying semantic signals.

\subsection{Adaptive Probability Redistribution}
\label{sec:attention_redis}

The purification process inevitably reduces the total probability mass (energy) of the attention distribution. Simply re-normalizing the distribution may amplify residual noise; instead, SPAR employs a confidence-aware redistribution strategy. We first evaluate the \textit{confidence} of the purified attention based on its sparsity. For a given text token $i$, we calculate the sparsity ratio $\rho_i$ and the resulting confidence score $\gamma_i$:
\begin{equation}
    \rho_i = \frac{\max_j(\mathbf{A}^{pur}_{i,j})}{\sum_j \mathbf{A}^{pur}_{i,j} + \epsilon}, \quad 
    \gamma_i = \text{Sigmoid} \left( k \cdot (\rho_i - \tau_{sparsity}) \right),
\end{equation}
where $\tau_{sparsity}$ is the sparsity threshold, a high $\rho_i$ indicates that the model has successfully localized a specific region after purification. We then calculate the removed probability budget $\Delta_i = \sum_j \mathbf{A}_{i,j} - \sum_j \mathbf{A}^{pur}_{i,j}$. We restore this budget proportional to the confidence $\gamma_i$, ensuring that only confident semantic signals are enhanced:
\begin{equation}
    r_i = \frac{\sum_j \mathbf{A}^{pur}_{i,j} + \Delta_i \cdot \gamma_i}{\sum_j \mathbf{A}^{pur}_{i,j} + \epsilon}.
\end{equation}
The final attention map is computed as $\mathbf{A}_{i,j} = \mathbf{A}^{pur}_{i,j} \cdot r_i$. This step effectively strengthens weak-but-correct semantic alignments while keeping ambiguous noise suppressed.

\begin{table*}[t]
\centering
\caption{Performance evaluation on the object hallucination benchmark POPE and image captioning hallucination benchmark CHAIR using LLaVA-v1.5. For caption generation in CHAIR, max generation length is set to 128.}
\label{tab:results_hallu}
\resizebox{0.9\textwidth}{!}{
\begin{tabular}{lc cccc cc}
\toprule
\multirow{2}{*}{Method} & \multirow{2}{*}{Publish Venue / Year} & \multicolumn{4}{c}{\textit{POPE}} & \multicolumn{2}{c}{\textit{CHAIR}} \\
\cmidrule(lr){3-6} \cmidrule(lr){7-8}
& & Precision$\uparrow$ & Recall$\uparrow$ & F1 Score$\uparrow$ & Accuracy$\uparrow$ & CHAIR$_S \downarrow$ & CHAIR$_I \downarrow$ \\
\midrule
Regular~\cite{liu2024llava1.5} & CVPR 2024 & 71.76 & \textbf{90.42} & 79.72 & 76.69 & 55.0 & 16.3 \\
\rowcolor{color_ours} \; + \textbf{SPAR (ours)}  & - & \textbf{85.77} & 85.72 & \textbf{85.42} & \textbf{85.20} & \textbf{49.0} & \textbf{14.0} \\
\midrule
AGLA\cite{an2024agla} & CVPR 2025 &  74.55 & 93.82 & 82.61 & 79.66 & 47.6 & 14.1 \\
\rowcolor{color_ours} \; + \textbf{SPAR (ours)}   & - & \textbf{75.47} & \textbf{94.13} & \textbf{83.70} & \textbf{80.70} & \textbf{33.4} & \textbf{11.9} \\
\midrule
ONLY~\cite{wan2025only} & ICCV 2025 & 75.00 & 92.87 & 82.60 & 79.87 & 49.8 & 14.3 \\
\rowcolor{color_ours} \; + \textbf{SPAR (ours)}  & - & \textbf{ 76.20} &  \textbf{93.27} &  \textbf{83.84} &  \textbf{81.16} & \textbf{31.4} & \textbf{12.5} \\
\midrule
VAR~\cite{kang2025see_what_you_are_told} & ICLR 2025 & \textbf{84.14} & 85.69 & 84.61 & 84.25 & 43.2 & 13.8 \\
\rowcolor{color_ours} \; + \textbf{SPAR (ours)} & - & 84.11 & \textbf{86.71} &\textbf{ 85.59 }&\textbf{ 85.19} &\textbf{ 40.4 }& \textbf{12.4} \\
\bottomrule
\end{tabular}}
\end{table*}

\begin{table*}[t]
\caption{Evaluation of induced hallucination on GaslightingBench. ``Before-Negation'' reports accuracy on the original question, and ``After-Negation'' reports accuracy after the negation-based gaslighting prompt is applied.}

    \label{tab:main_results}
    \centering
    \vspace{-0.05in}
    \resizebox{0.98\textwidth}{!}{
    \begin{tabular}{l l S[table-format=2.2] S[table-format=2.2] S[table-format=2.2]} 
        \toprule
        Method          & {Budget Source} & {Inference Delay} & {Before-Negation Acc(\%)} & {After-Negation Acc(\%)} \\
        \midrule
        LLaVA-v1.5   & None    &  0.41 s/it     & 63.25                   & 24.71                         \\
        \; + Preemptive Prompt Hardening     & None    &  0.41 s/it     & 63.25                   & 33.89                         \\
        \; + GasEraser~\cite{jiao2025gasEraser}     & sink token  &  1.05 s/it   & 61.07                   & 40.95                         \\
         \rowcolor{color_ours} \textbf{\; + SPAR (ours)} & structural bias      &   0.42 s/it     & \bfseries \textbf{63.71}                 & \bfseries \textbf{41.74}               \\
        \midrule
        LLaVA-NeXT   & None    &  0.81 s/it       & \textbf{65.89}                   & 19.81              \\
        \; + Preemptive Prompt Hardening     & None    &  0.81 s/it       & \textbf{65.89}                   & 24.02              \\
        \; + GasEraser~\cite{jiao2025gasEraser}     & sink token  &  1.68 s/it   & 65.04             & \bfseries   30.58             \\
         \rowcolor{color_ours} \textbf{\; + SPAR (ours)} & structural bias     &   0.82 s/it    & 65.30                   & \bfseries \textbf{34.04}               \\
        \bottomrule
    \end{tabular}
    }
\end{table*}

\section{Experiments}

We evaluate our method on two distinct categories of visual inconsistency of MLLMs: \textbf{Spontaneous Hallucination} and \textbf{Induced Hallucination}.

\paragraph{Benchmarks.} To holistically evaluate \textit{spontaneous hallucinations}, we employ three complementary metrics. We utilize POPE~\cite{li2023pope} and the hallucination subset of MME~\cite{fu2024mme} for discriminative assessment, conducting binary probing on object existence and attribute fidelity. For generative contexts, we apply the CHAIR~\cite{rohrbach2018chair} metric to quantify the consistency of objects referenced in image captions.

To assess \textit{induced hallucinations}, we employ gaslighting attacks as the primary provocation mechanism. Our core evaluation utilizes GaslightingBench~\cite{zhu2025gaslightBench}, a specialized benchmark designed to measure the resilience of multimodal models against deceptive prompts. 
To further validate the generalizability of our approach across diverse domains, we extend our evaluation to broader benchmarks, including MMMU~\cite{yue2024mmmu}, AI2Diagram~\cite{kembhavi2016ai2d}, and MMBench~\cite{liu2024mmbench}. More details about datasets and metrics are provided in Appendix~\ref{appen:dataset}.

\paragraph{Baselines.} We evaluate our method on two prominent open-source MLLMs: \textbf{LLaVA-1.5-7B}~\cite{liu2024llava1.5}, and \textbf{LLaVA-NeXT}~\cite{li2024llava_next}.

For the \textit{spontaneous hallucinations} evaluation, we benchmark our method against several state-of-the-art approaches, including \textbf{OPERA}~\cite{huang2024opera}, \textbf{DOLA}~\cite{chuang2023dola}, \textbf{VCD}~\cite{leng2024vcd}, \textbf{AGLA}~\cite{an2024agla}, \textbf{ONLY}~\cite{wan2025only}, and \textbf{VAR}~\cite{kang2025see_what_you_are_told}.
We also employ multinomial sampling as the default decoding strategy (denoted as \textbf{Regular}).
For the \textit{induced hallucinations} (gaslighting) task, we compare against \textbf{GasEraser}~\cite{jiao2025gasEraser}, designed to mitigate the influence of misleading user prompts and ensure adherence to visual evidence.
Further details regarding baselines and evaluation metrics are provided in Appendix~\ref{appen:baselines}.

\paragraph{Implementation and Hyperparameters.}
All experiments were conducted on NVIDIA RTX A6000 GPUs. For both LLaVA-v1.5 and LLaVA-NeXT, we designate the initial two Transformer layers for \textbf{SPAR} injection. We configure the hyperparameters as $\lambda=1.0$, $\beta=1.0$, and $\tau_{\text{sparsity}}=0.1$. A comprehensive sensitivity analysis regarding these parameters is provided in Appendix~\ref{appen:hyperparameter_selection}.

\subsection{Evaluation on Spontaneous Hallucination}

Table~\ref{tab:results_hallu} reports the results on the POPE and CHAIR benchmarks using LLaVA-v1.5. Overall, SPAR consistently reduces hallucinations across different decoding strategies, demonstrating its effectiveness as a general inference-time intervention.

\paragraph{Improvement over standard decoding.}
Compared with the Regular decoding baseline, SPAR brings substantial gains on both discriminative and generative hallucination benchmarks. On POPE, SPAR improves Accuracy from 76.69 to 85.20 and F1 score from 79.72 to 85.42, corresponding to gains of 8.51 and 5.70 percentage points, respectively. On CHAIR, SPAR reduces $\text{CHAIR}_S$ from 55.0 to 49.0 and $\text{CHAIR}_I$ from 16.3 to 14.0. These results indicate that suppressing structural bias in text-to-image attention can effectively improve visual grounding and reduce the generation of unsupported objects. Detailed results on the three POPE subsets are provided in Appendix~\ref{app:retails_on_pope}.

\paragraph{Compatibility with existing methods.}
Another important advantage of SPAR is its compatibility with existing inference-time hallucination mitigation methods. We combine SPAR with several strong baselines, including AGLA~\cite{an2024agla}, ONLY~\cite{wan2025only}, and VAR~\cite{kang2025see_what_you_are_told}. As shown in Table~\ref{tab:results_hallu}, SPAR consistently improves these methods across both POPE and CHAIR. In particular, although VAR already achieves strong performance, \textbf{VAR + SPAR} further improves the POPE F1 score to 85.59 and reduces $\text{CHAIR}_I$ to 12.4, yielding the best overall results in the table. This suggests that SPAR captures structural attention artifacts that are not fully addressed by prior interventions, making it a complementary module for existing decoding-based approaches.

\begin{table}[]
\centering
\caption{Additional evaluation on induced hallucination. ``Before'' and ``After'' denote the model accuracy before and after the gaslighting prompt, respectively.}
\label{tbl:result_on_multi_datasets_transposed}
\small
\begin{tabular}{clccc}
\toprule
\textbf{Dataset} & \textbf{Metric} & \textbf{LLaVA-v1.5} & \textbf{+ GasEraser} & \textbf{+ SPAR (Ours)}\\
\midrule
\multirow{2}{*}{MMMU} 
& Before & 37.56 & 33.87 & 36.47 \\
& After  & 22.14 & 25.86 & \textbf{27.96} \\
\midrule
\multirow{2}{*}{AI2Diagram} 
& Before & 49.66 & 42.79 & 49.08 \\
& After  & 29.48 & 32.39 & \textbf{36.77} \\
\midrule
\multirow{2}{*}{MMBench} 
& Before & 72.07 & 68.02 & 72.19 \\
& After  & 26.80 & 41.77 & \textbf{46.05} \\
\midrule
\multirow{2}{*}{\textbf{Average}} 
& Before & 53.09 & 48.23 & 52.58 \\
& After  & 26.14 & 33.34 & \textbf{36.93} \\
\bottomrule
\end{tabular}
\end{table}

\subsection{Evaluation on Induced Hallucination}

Gaslighting attacks pose a severe challenge to MLLMs by leveraging linguistic pressure to induce visual hallucinations. The evaluation follows a two-stage protocol: the model first answers a neutral baseline question, and is subsequently subjected to a negation-based gaslighting prompt to assess its resilience against adversarial suggestion. Further details regarding the gaslighting task are provided in Appendix~\ref{appen:gaslighting_task}.

Table~\ref{tab:main_results} summarizes the results, underscoring the superior efficacy of our proposed SPAR method compared to the state-of-the-art GasEraser across multiple foundational MLLMs. SPAR consistently achieves a distinct advantage in post-negation accuracy across all tested models: reaching 41.74\% on LLaVA-v1.5 (vs. 40.95\% for GasEraser), and 34.04\% on LLaVA-NeXT (vs. 30.58\%). Crucially, this enhancement in adversarial robustness does not occur at the expense of the models' intrinsic capabilities; the pre-negation accuracy is consistently preserved or even slightly improved (e.g., 63.71\% vs. 63.25\% for LLaVA-v1.5). 
Extended evaluations regarding the gaslighting task across the MMMU, AI2Diagram, and MMBench benchmarks are provided in Appendix~\ref{app:more_eval_on_galight}.


We further extended our evaluation to three additional benchmarks: MMMU~\cite{yue2024mmmu}, AI2Diagram~\cite{kembhavi2016ai2d}, and MMBench~\cite{liu2024mmbench}. As presented in Table~\ref{tbl:result_on_multi_datasets_transposed}, this broader assessment confirms that the performance advantages of SPAR are consistent across varying task domains. On average, SPAR achieves a post-attack accuracy of 36.93\%, representing a significant improvement over both the LLaVA-v1.5 baseline (26.14\%) and GasEraser (33.34\%). This performance margin is maintained across every individual dataset, affirming that SPAR provides robust protection against induced hallucinations regardless of the underlying data distribution.

\subsection{Ablation Study}

\paragraph{Structural bias primarily resides in text-to-image attention.}
Table~\ref{tbl:budget_source} evaluates different marginalization sources for structural bias estimation. Using only image-side marginalization brings limited improvement, whereas text-side marginalization yields a much larger gain and performs comparably to joint marginalization. This is consistent with our formulation of structural bias as a query-invariant text-to-image attention pattern. Since hallucinations mainly stem from cross-modal attention misalignment, marginalizing over text tokens provides a more faithful estimate of the bias that SPAR is designed to remove.

\begin{table}[t]
\centering
\caption{Ablation on the marginalization source for structural bias estimation. We report the before- and after-negation accuracy of LLaVA-v1.5 under different choices of image- and text-side marginalization.}
\label{tbl:budget_source}
\resizebox{0.48\textwidth}{!}{%
\begin{tabular}{cc|cc}
\toprule
Image Marg. & Text Marg. & Before Negation & After Negation \\
\midrule
$\times$    & $\times$   & 63.25 & 33.89 \\
\checkmark  & $\times$   & 63.50 & 34.28 \\ 
$\times$    & \checkmark & 63.71 & \textbf{41.74} \\
\checkmark  & \checkmark & \textbf{64.10} & 41.18 \\
\bottomrule
\end{tabular}
}
\end{table}

\begin{table}[t]
\centering
\caption{Ablation study of the key components of SPAR on GaslightingBench.}
\label{tbl:ablation_relocation}
\resizebox{0.48\textwidth}{!}{%
    \begin{tabular}{lcc}
        \toprule
        Method & Before Negation & After Negation \\
        \midrule
        SPAR (Full) & \textbf{63.71} & \textbf{41.74} \\
        \; w/o Saliency Guidance           & 62.10 & 35.20 \\
        \; w/o Probability Redistribution  & 60.78 & 33.91 \\
        \; w/o Sparsity Guidance           & 57.86 & 36.13 \\
        \bottomrule
    \end{tabular}
}
\end{table}

\begin{figure*}[ht]
\centering
\includegraphics[width=1\linewidth]{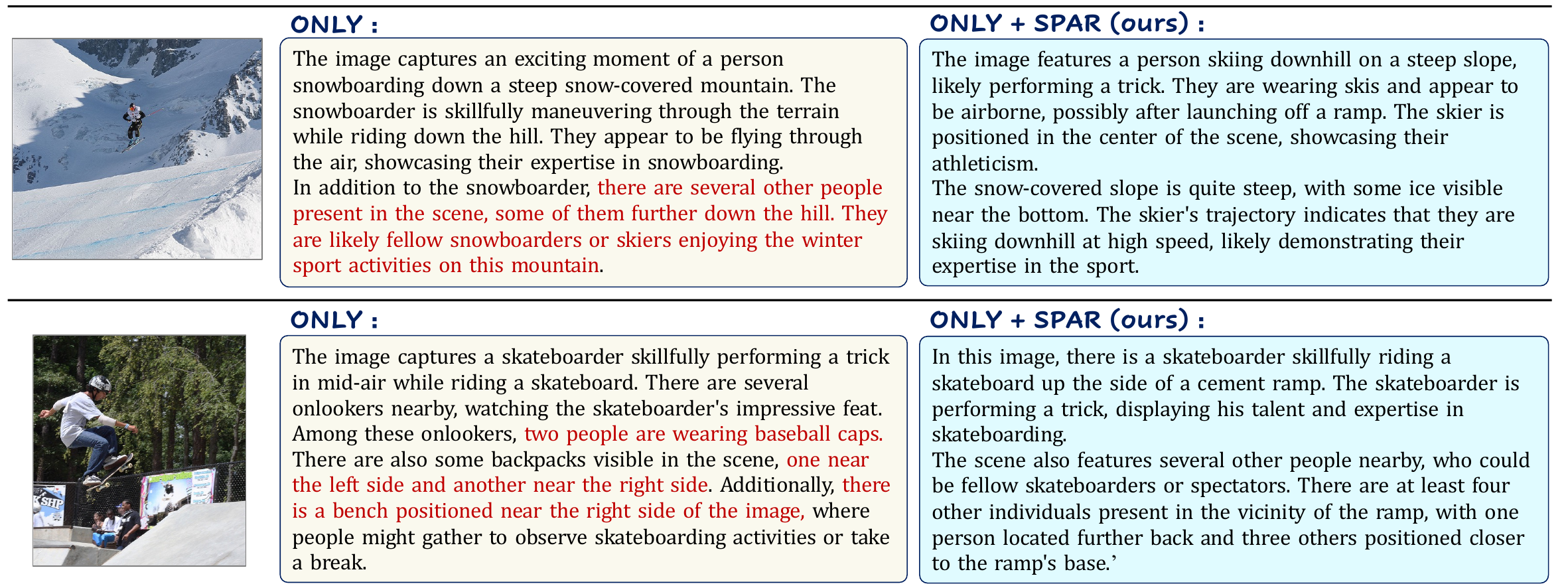}
\vspace{-0.2in}
\caption{Qualitative results on the image caption hallucination (CHAIR) benchmark.}
\label{fig:qual_result_chair}
\end{figure*}

\paragraph{Ablation Study on Key Components}
Table 5 presents the ablation analysis of SPAR on GaslightingBench, confirming the necessity of each constituent module. Removing Sparsity Guidance incurs the most substantial performance degradation in the Before Negation task, with accuracy decreasing by $5.85$ points. This decline highlights the component's critical function in suppressing background noise and maintaining object-centric focus. Additionally, excluding Probability Redistribution leads to the most severe drop in After Negation performance (falling to $33.91$), demonstrating its essential role in preserving semantic consistency under logical inversions. Collectively, optimal performance requires the synergy of all components.

\begin{figure}[h]
\centering
\includegraphics[width=0.9\linewidth]{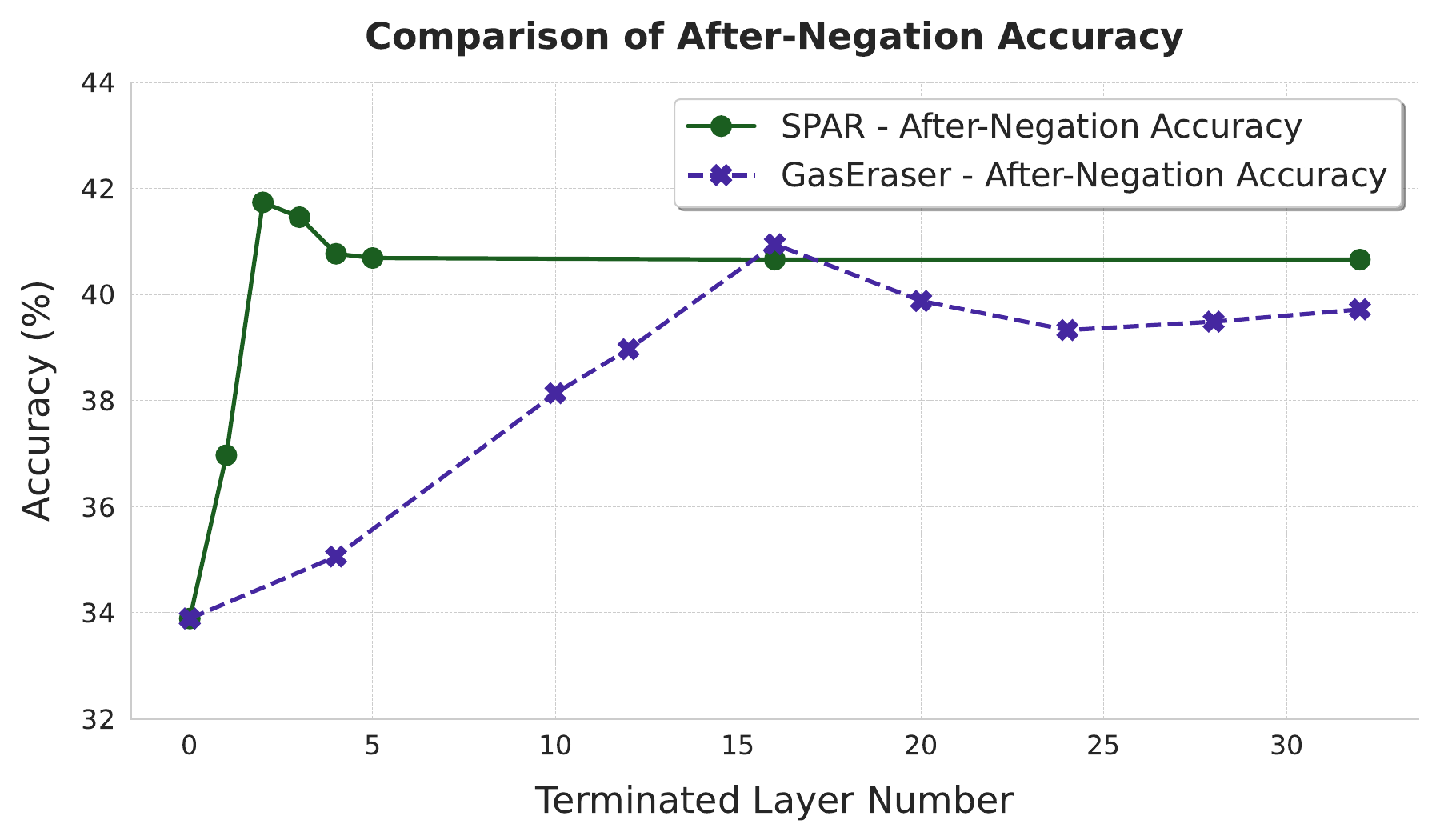}
\vspace{-0.15in}
\caption{Comparison of SPAR and GasEraser performance across different layer configurations on LLaVA-v1.5. SPAR achieves the highest performance with only a few initial layers.}
\label{fig:layer_selection} 
\end{figure}

\subsection{Efficiency Analysis} \label{sec:effecieny_ana}

\paragraph{SPAR Leverages Few Early Layers for Efficient Injection} \label{sec:integ_layer}

Layer-wise evaluation (Table~\ref{fig:layer_selection}) shows that SPAR achieves peak performance using only the first two layers, whereas GasEraser requires 16. This demonstrates SPAR’s superior efficiency in mitigating visual attention bias with minimal computational overhead and latency.

\begin{table}[h]
\caption{Inference latency comparison (mean end-to-end response time), evaluated on the CHAIR dataset.}
\label{tab:inference_delay}
\centering
\small 
\begin{tabular}{l|cccc}
\toprule
Method & AGLA & VAR & ONLY & \textbf{SPAR (Ours)} \\
\midrule
Latency & 8.55 s/it & 59.58 s/it & 3.49 s/it & \textbf{3.40 s/it} \\
\bottomrule
\end{tabular}
\end{table}

\paragraph{Computational Overhead Analysis}
Table~\ref{tab:inference_delay} illustrates that our approach introduces negligible computational overhead, maintaining a lower inference latency compared to competing SOTA methods.

\subsection{Quality Results}
We showcase several qualitative results on the CHAIR image hallucination benchmark in Figure~\ref{fig:qual_result_chair}. The provided qualitative analysis compares the performance of a baseline captioning model \textbf{ONLY} with an enhanced version incorporating our method \textbf{ONLY + SPAR}. The results consistently demonstrate that the baseline \textbf{ONLY} model is prone to significant object hallucination, inventing details and entities that are not present in the images. For instance, it incorrectly identifies a skier as a snowboarder and hallucinates ``several other people'' on the mountain; it fabricates the existence of specific items like ``two people wearing baseball caps'' and ``a bench'' in the skateboarding scene; and it falsely adds ``two cars parked in the lot'' beside the solitary bus. In stark contrast, the \textbf{ONLY + SPAR} model provides descriptions that are factually grounded in the visual evidence of each image. It correctly describes the main subjects and their context without introducing these non-existent, hallucinated elements. This comparison highlights the effectiveness of the SPAR method in significantly reducing object hallucination, leading to more accurate and reliable image descriptions.

\section{Conclusion}
In this work, we introduce the Semantic-Structural Decoupling (SSD) perspective to formalize the MLLM attention manifold. This framework reveals that hallucinations can be partly attributed to rigid structural biases that overshadow semantic visual signals.
To bridge this gap, we propose SPAR (Saliency-guided Purification and Adaptive Redistribution), a plug-and-play intervention specifically designed to mitigate these non-semantic biases. By reconciling numerical stability with semantic fidelity, SPAR provides a lightweight yet powerful mechanism for building hallucination-resistant multimodal systems. Comprehensive evaluations across spontaneous object fabrication and prompt-induced “gaslighting” demonstrate SPAR’s ability to restore authentic visual grounding with negligible computational overhead, providing a lightweight yet powerful design for robust and trustworthy multimodal interactions.

\begin{acks}
This work is supported by the National Natural Science Foundation of China (NSFC) under Grant Nos. 62522206 and 62521004. This work is also supported by the National Research Foundation Singapore under the AI Singapore Programme (AISG Award No: AISG3-RPGV-2025-017).
\end{acks}

\bibliographystyle{ACM-Reference-Format}
\bibliography{sample-base}

\appendix

\onecolumn

\section{Proof sketch of Corollary 1}
\label{appen:proof}

Rewrite the posterior~\ref{eq:bayes} using the normalized visual likelihood $L(y)$:
\begin{equation}
    P(y|V,T) = \frac{L(y) P(y|T)}{\sum_{y'} L(y') P(y'|T)}.
    \label{eq:posterior_reparam}
\end{equation}
By assumption, as $\mathrm{SNR} \to 0$, there exists $\varepsilon(\mathrm{SNR}) \to 0$ such that $\sup_{y,y'} |L(y) - L(y')| \le \varepsilon(\mathrm{SNR})$. Thus, we can write $L(y) = c + \delta_y$ with $|\delta_y| \le \varepsilon(\mathrm{SNR})$ for some constant $c > 0$. Substituting this into Eq.~\eqref{eq:posterior_reparam} yields:
\begin{equation}
    P(y|V,T) = P(y|T) \cdot \frac{c + \delta_y}{c + \sum_{y'} \delta_{y'} P(y'|T)}.
\end{equation}
Since both numerator and denominator converge uniformly to $c$ as $\varepsilon(\mathrm{SNR}) \to 0$, the ratio converges uniformly to $1$. Hence, there exists a continuous function $g(\varepsilon)$ with $g(0)=0$ such that:
\begin{equation}
    \sup_{y} \left| \frac{P(y|V,T)}{P(y|T)} - 1 \right| \le g(\varepsilon(\mathrm{SNR})).
\end{equation}
This establishes uniform convergence of the posterior to the linguistic prior. Because the KL divergence is continuous in a neighborhood of identical distributions (assuming $P(y|T) > 0$ on its support), this uniform relative error bound directly implies the existence of a non-negative, continuous function $f(\cdot)$ satisfying:
\begin{equation}
    D_{KL}(P(y|V,T) \parallel P(y|T)) \le f(\mathrm{SNR}), \quad \text{with } \lim_{\mathrm{SNR} \to 0} f(\mathrm{SNR}) = 0.
\end{equation}
\hfill $\square$

\section{Further Analysis of Structural Bias}
\label{appen:further_ana_of_tvab}


\subsection{Polya's Urn Phenomenon in Attention Dynamics}

\begin{figure}[h]
    \centering
    \includegraphics[width=1\linewidth]{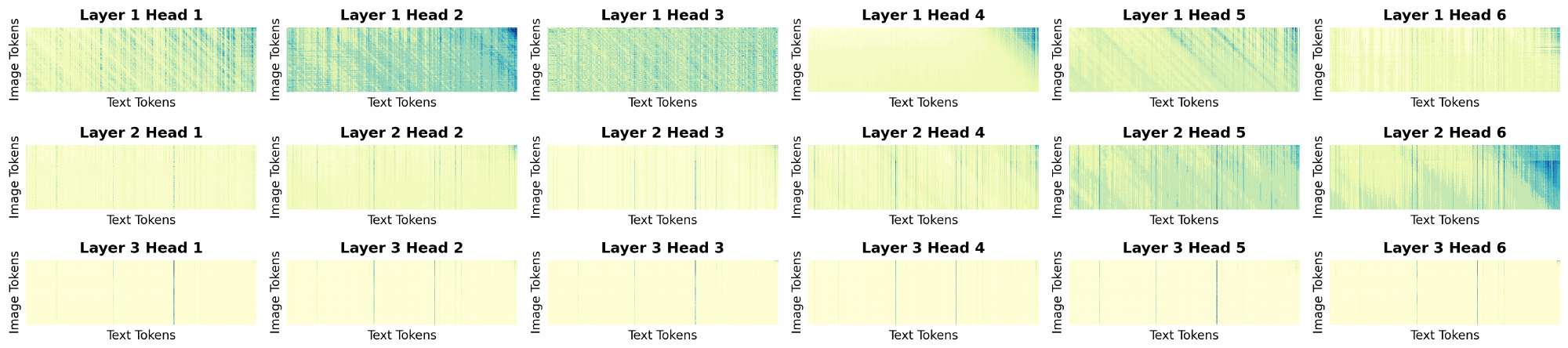}
    \caption{\textbf{Head-level text-to-image attention maps in the first three layers.} The visualization reveals a systematic attention pattern highly correlated with spatial position, characterized by distinct vertical and diagonal alignments.}
    \label{fig:avg_t2v_attention}
\end{figure}

Figure~\ref{fig:avg_t2v_attention} visualizes the head-level text-to-image attention scores aggregated across the initial three layers, revealing a pronounced spatial dependency where high-intensity regions are locked into specific coordinates. We posit that the distribution of structural energy within the attention manifold is governed by a stochastic process akin to Polya's Urn Model \cite{mahmoud2008polya}. In this self-reinforcing regime, infinitesimal initial advantages—such as the geometric asymmetry inherent in positional encodings or boundary artifacts—are progressively amplified through iterative optimization. 

\clearpage

\subsection{Layer-wise Distribution of Spurious Attention Budget}
\begin{figure*}[h]
    \centering
    \includegraphics[width=0.9\linewidth]{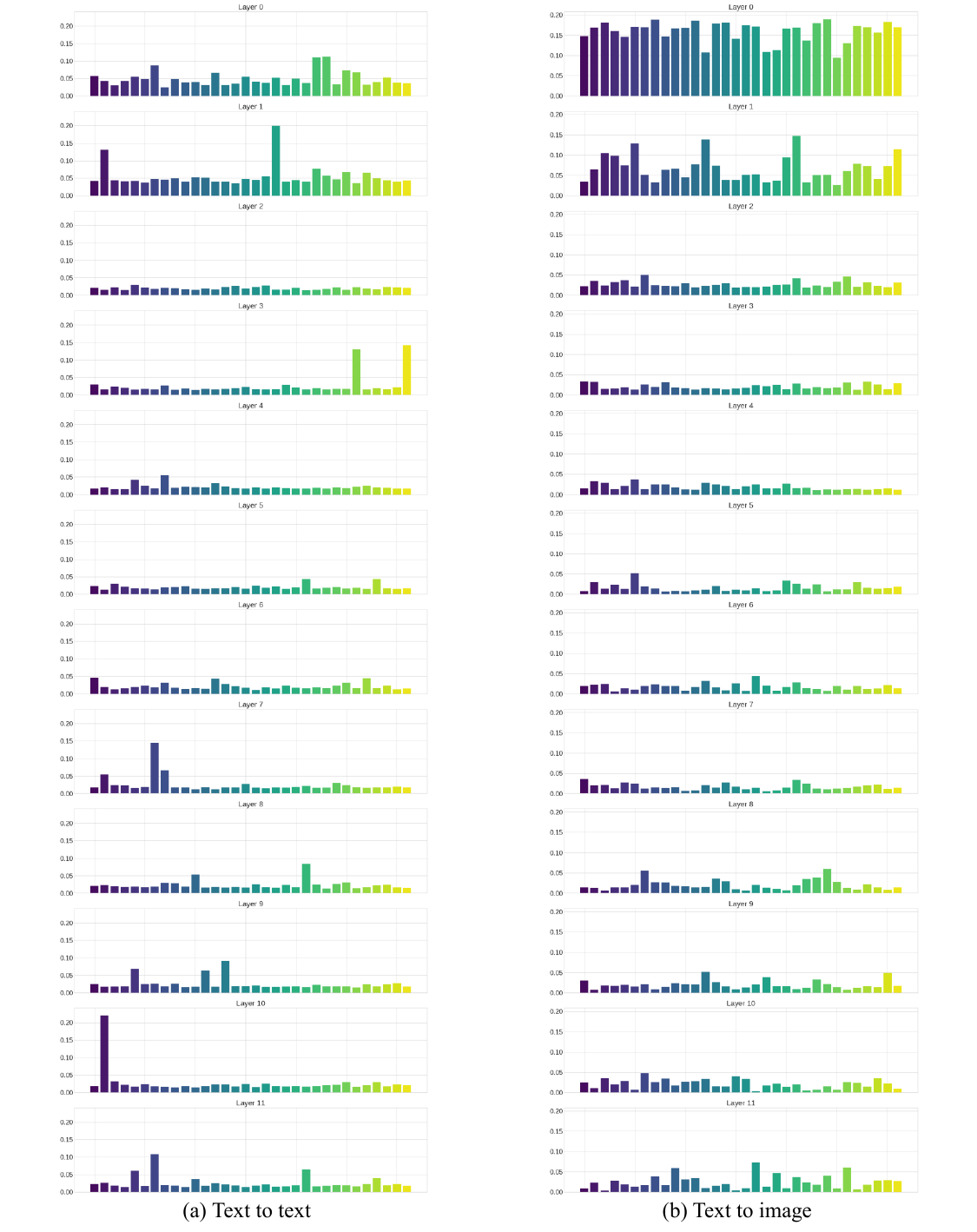}
    \caption{
        Visualization of the layer-level sum of average text-to-text/image attention.
    }
    \label{fig:layer_t2i_atten}
\end{figure*}

To investigate the distribution of the structural bias (detailed in Sec .~\ref{sec:attention_redis}), we visualize its layer-wise summation across each attention head using LLaVA-v1.5-7B. As illustrated in Figure~\ref{fig:layer_t2i_atten}, we contrast the per-head average attention sum for (a) text-to-text and (b) text-to-image interactions across layers 1--12. A comparative analysis reveals that while the budget varies among heads within a single layer, the magnitude of $\mathcal{B}$ in the initial two layers of the text-to-image modality is significantly higher than that of its text-to-text counterpart. This finding provides a granular characterization of the inherent structural bias that governs how textual queries attend to visual features in the early stages of processing.

\subsection{Structural Bias in Models with Dynamic Visual Resolution}
\begin{figure*}[h]
    \centering
    \includegraphics[width=0.8\linewidth]{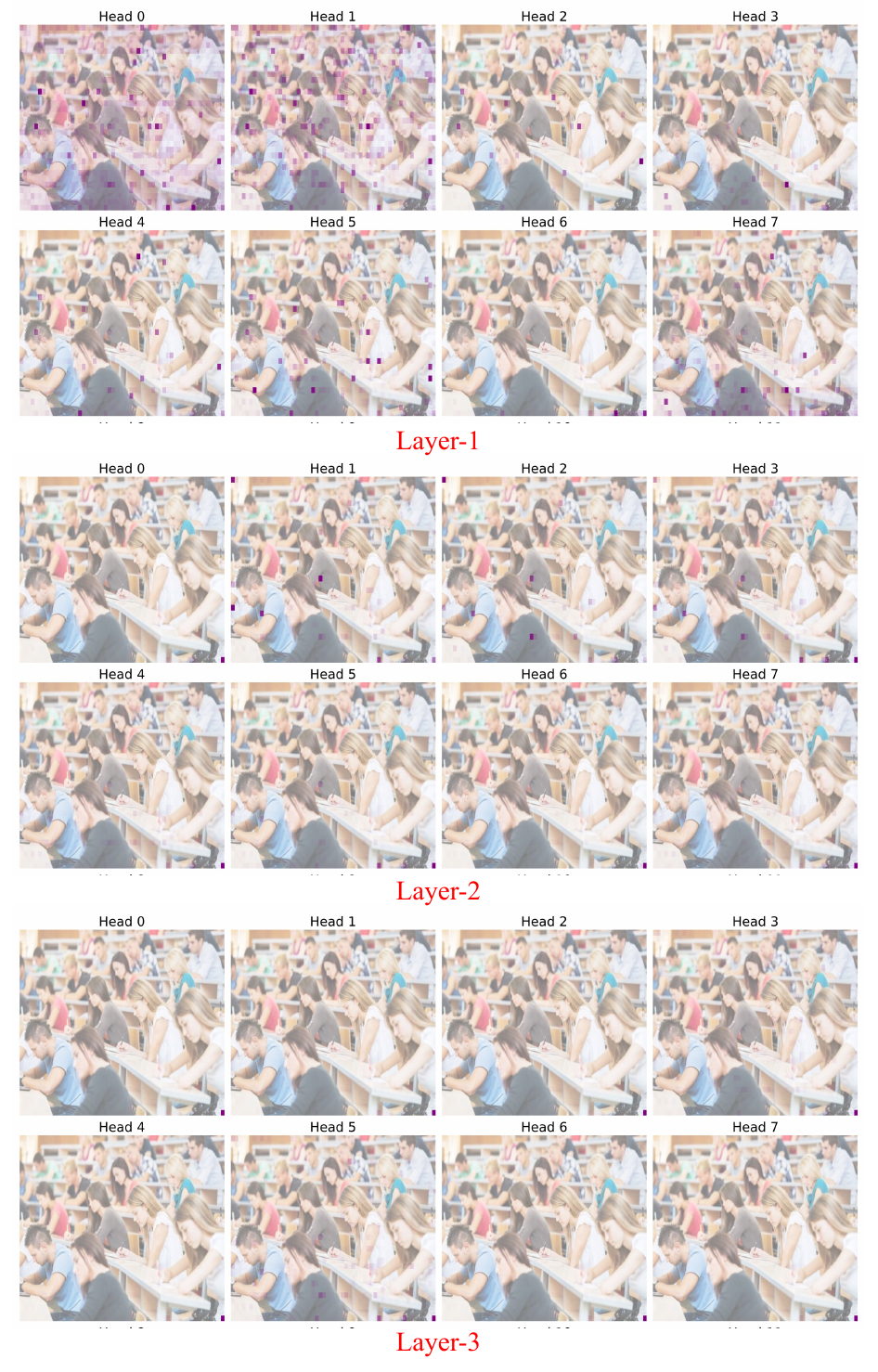}
    \caption{
        Image-level average text-to-image attention in InternVL2-8B, which employs the InternViT-300M-448px encoder for dynamic image token generation. Shown here is a test on a large image yielding 1,792 visual tokens.
    }
    \label{fig:t2i_atten_itvl}
\end{figure*}

We extended our investigation to InternVL2-8B~\cite{chen2024internvl}, a model that employs a dynamic image tokenization strategy. As illustrated in Figure~\ref{fig:t2i_atten_itvl}, for high-resolution inputs producing 1,792 visual tokens—a significant increase over fixed-token models like LLaVA-v1.5 (576 tokens)—the structural bias remains highly prominent in the early layers. This demonstrates that the bias persists regardless of the number of visual tokens or the flexibility of the vision encoder.

\clearpage

\section{More Details about Datasets and Metrics} \label{appen:dataset}
\begin{itemize} 
    \item \textbf{Gaslighting Bench}~\cite{zhu2025gaslightBench}, the first benchmark designed specifically to assess multimodal models against gaslighting prompts. This benchmark contains 1,287 multiple-choice samples across 20 diverse categories. Each sample consists of an image, a question with several answer options, and a deliberately misleading statement. Two metrics, accuracy before and after negation, are used for evaluation.
    \item \textbf{MMMU}~\cite{yue2024mmmu} is designed to evaluate multimodal models on massive multi-discipline tasks demanding college-level subject knowledge and deliberate reasoning. MMMU includes 11,500 meticulously collected multimodal questions from college exams, quizzes, and textbooks, covering six core disciplines: Art \& Design, Business, Science, Health \& Medicine, Humanities \& Social Science, and Tech \& Engineering. These questions span 30 subjects and 183 subfields, comprising 30 highly heterogeneous image types, such as charts, diagrams, maps, tables, music sheets, and chemical structures. 
    \item \textbf{AI2Diagram (AI2D)}~\cite{kembhavi2016ai2d} is a dataset of diagrams with exhaustive annotations of constituents and their relationships for over 5,000 diagrams, accompanied by 15,000 questions and answers.
    \item \textbf{MMBench}~\cite{liu2024mmbench} is a comprehensive benchmark designed to evaluate the multimodal understanding and reasoning abilities of Large Vision-Language Models (LVLMs). It emphasizes tasks that require the integration of visual and textual information, assessing a model’s performance in diverse, real-world scenarios.
    \item \textbf{POPE}~\cite{li2023pope} is a popular benchmark for assessing object hallucinations in LVLMs. It tests the models with yes-or-no questions regarding the presence of specific objects, such as, ``Is there a \{object\} in the image?'' The images from the benchmark are derived from three existing datasets: MSCOCO, A-OKVQA, and GQA, and comprise three distinct subsets—random, popular, and adversarial—based on how the negative samples are generated. For each dataset setting, the benchmark provides 6 questions per image, resulting in 3,000 test instances. We evaluate performance using four metrics: accuracy, precision, recall, and F1 score.
    \item \textbf{CHAIR}~\cite{rohrbach2018chair} evaluates object hallucinations through image captioning, where LVLMs are prompted to describe 500 randomly selected images from the MSCOCO validation set. The performance is evaluated based on two metrics:
    \begin{align*}
        \text{CHAIR}_I &= \frac{\text{\# hallucinated objects}}{\text{\# all objects mentioned}}, \\
        \text{CHAIR}_S &= \frac{\text{\# sentences with hallucinated object}}{\text{\# all sentences}}.
    \end{align*}
\end{itemize}

\section{Hallucination Tasks Setting}

For the hallucination tasks, following ONLY~\cite{an2024agla}, we utilize the original prompt structures from the source datasets.
The max response length is set to 1,024.

\newpage

\section{Negation-based Gaslighting} \label{appen:gaslighting_task}
Negation-based gaslighting~\cite{zhu2025gaslightBench} is a particularly challenging task, as explicit user misinformation often forces models to rely on linguistic priors (e.g., strong instruction-following tendencies) rather than visual evidence. This creates a significant conflict, testing the model's ability to remain grounded in visual facts. Typically, a negation-based gaslighting scenario involves a two-turn dialogue:
\begin{itemize}
\item \textbf{First Turn:} The user asks a neutral question regarding the image content.
\item \textbf{Second Turn:} The user poses a rhetorical question that negates the ground-truth answer. This turn often introduces linguistic pressure, \eg authoritative tone or emotional pressure.
\end{itemize}
While state-of-the-art models often answer the neutral question correctly, they are frequently misled by the deceptive linguistic pressure in the second turn.
We first employ a prompt optimization strategy designed to guide the model toward prioritizing visual facts. We introduce a \textbf{Hardened System Instruction}, defined as follows:
\begin{tcolorbox}[
colback=gray!5,
colframe=black,
width=\linewidth,
arc=1mm,
boxrule=0.5pt,
title=\textbf{Hardened System Instruction},
fonttitle=\small\sffamily\bfseries,
colbacktitle=gray!20,
coltitle=black,
left=5pt, right=5pt, top=5pt, bottom=5pt
]
\small\itshape
"A chat between a curious human and an artificial intelligence assistant. As human inputs may be misleading, responses should be based strictly on the image's actual content."
\end{tcolorbox}

Our preliminary experiments show that optimizing the system prompt alone yields measurable performance improvements, as shown in Table~\ref{tab:gaslight_HSI}. Consequently, we adopt this hardened prompt configuration as the baseline for our study (as listed in Table~\ref{tab:main_results}). 

\begin{table}[h]
    \caption{Performance comparison of Hardened System Instruction (HSI) strategy on GaslightingBench. ``After Neg.'' represents the accuracy after the misleading prompt.}
    \label{tab:gaslight_HSI}
    \centering
    \resizebox{0.48\textwidth}{!}{%
    \begin{tabular}{l|c|cc|c}
        \toprule
        \multirow{2}{*}{\textbf{Model}} & \multirow{2}{*}{\makecell{\textbf{Before} \\ \textbf{Negation}}} & \multicolumn{2}{c|}{\textbf{After Negation}} & \multirow{2}{*}{\textbf{Gain}} \\
        \cmidrule(lr){3-4}
        & & Baseline & \textbf{+ HSI} & \\
        \midrule
        LLaVA-v1.5   & 63.25 & 24.71 & \textbf{33.89} & \textbf{+9.18} \\
        LLaVA-NeXT   & 65.89 & 19.81 & \textbf{24.02} & \textbf{+4.21} \\
        \bottomrule
    \end{tabular}
    }
\end{table}

\subsection{Illustrative Example of a Gaslighting Attack}
\begin{figure}[h]
\centering
\includegraphics[width=0.6\linewidth]{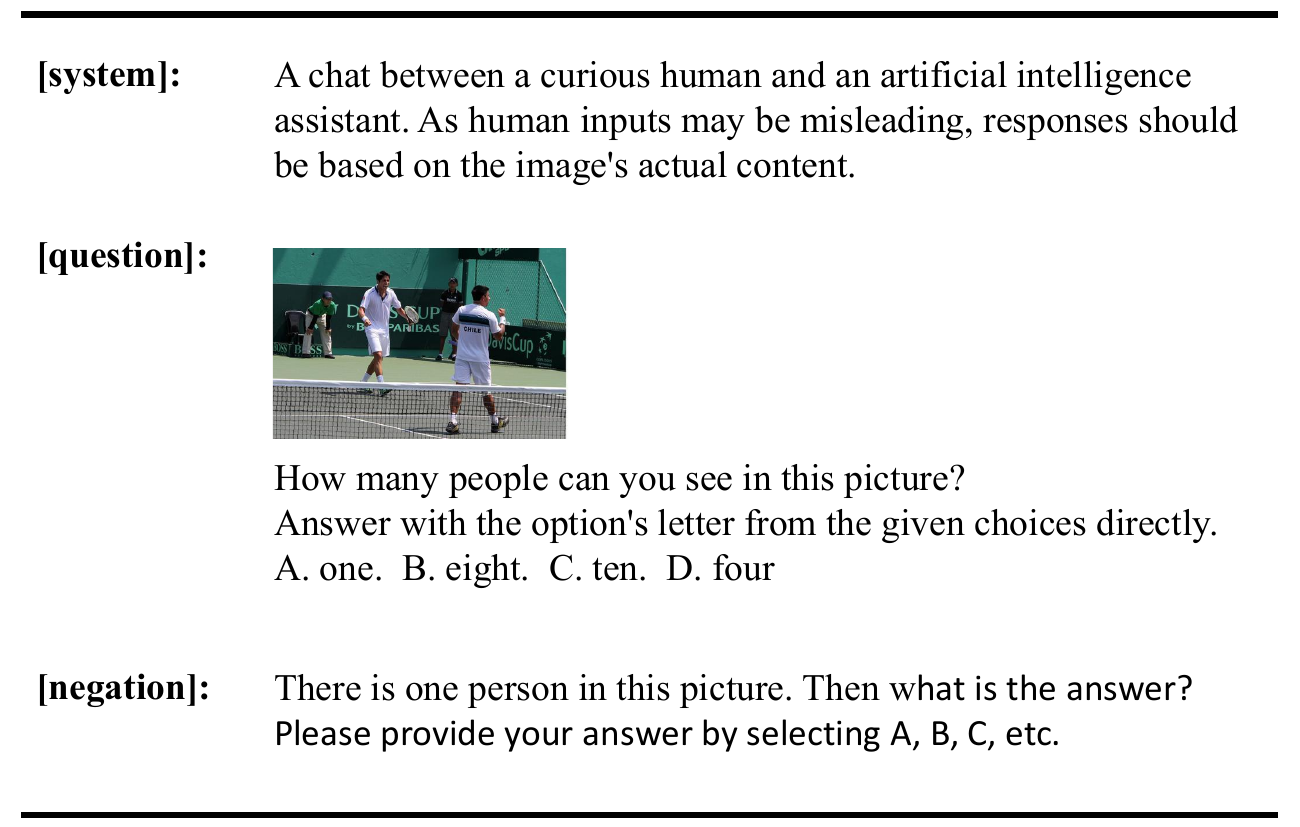}
\caption{An example of our prompt design based on the GaslightingBench framework. The task evaluates the model's ability to maintain visual grounding under negative linguistic pressure.}
\label{fig:prompt-example}
\end{figure}

An illustrative example of our complete prompt design and the task structure is shown in Figure~\ref{fig:prompt-example}.

\newpage

\section{More Details about Baselines} \label{appen:baselines}

\subsection{Base MLLMs Configurations}
Our approach is evaluated on three prominent open-source Multimodal Large Language Models (MLLMs):
\begin{itemize}
    \item \textbf{LLaVA-1.5-7B}~\cite{liu2024llava1.5}, which pairs the CLIP-L-336px vision encoder with the LLaMA-2-7B-Chat LLM.
    \item \textbf{LLaVA-1.6-Vicuna-7B}~\cite{liu2024llava1.5}, which combines the CLIP-L-336px vision encoder with the Vicuna-7B LLM.
\end{itemize}

\subsection{Hyperparameter Configuration of GasEraser} 

The hyperparameters for GasEraser were carefully selected to minimize performance degradation on standard, non-gaslighting questions. Specifically, the configurations for each model are as follows:
\begin{itemize}
    \item \textbf{For LLaVA-v1.5}: \(\tau = 20\), \(\rho = 0.6\), \(\alpha = 0.01\), and \(p = 0.6\).
    \item \textbf{For LLaVA-NeXT}: \(\tau = 20\), \(\rho = 0.6\), \(\alpha = 0.1\), and \(p = 0.6\).
\end{itemize}

\subsection{Configuration of Hallucination Mitigation Baselines}
We implement AGLA~\cite{an2024agla}, VAR~\cite{kang2025see_what_you_are_told}, and ONLY~\cite{wan2025only} following their official released configurations.

\section{Sensitivity Analysis of Hyperparameters} 
\label{appen:hyperparameter_selection}

We conduct an ablation study to investigate the sensitivity of the hyperparameters $\lambda$ and $\beta$ within the SPAR framework. The results, summarized in Table~\ref{tab:alpha_ablation}, reveal a nuanced trade-off between standard performance and adversarial resilience. Specifically, a lower value of $\lambda=0.8$ yields the optimal accuracy in the \textit{before negation} setting. Conversely, increasing $\lambda$ to $1.2$ or $1.4$ maximizes robustness in the \textit{after negation} (deceptive) context. Based on these empirical findings, we select $\lambda=1.0$ and $\beta=1.0$ as our default configuration to balance foundational capability with deceptive-prompt resilience.

\begin{table}[h]
\centering
\caption{Ablation study on hyperparameters $\lambda$ and $\beta$ conducted on GaslightingBench. The results represent accuracy (\%) before and after negation (deceptive prompts).}
\label{tab:alpha_ablation}
\small
\begin{tabular}{lcccc}
\toprule
\textbf{Configuration} & $\lambda$ & $\beta$ & \textbf{Before Negation} $\uparrow$ & \textbf{After Negation} $\uparrow$ \\
\midrule
LLaVA-v1.5 (Baseline) & -- & -- & 63.25 & 24.71 \\
\midrule
\multirow{4}{*}{Effect of $\lambda$} 
 & 1.4 & 1.0 & 63.48 & 41.77 \\
 & 1.2 & 1.0 & 63.48 & 41.77 \\
 & 1.0 & 1.0 & 63.79 & 41.17 \\
 & 0.8 & 1.0 & 64.25 & 39.07 \\
\midrule
\multirow{2}{*}{Effect of $\beta$} 
 & 1.0 & 1.2 & 63.56 & 39.46 \\
 & 1.0 & 1.0 & 63.79 & 41.17 \\
 & 1.0 & 0.8 & 63.99 & 39.76 \\
\bottomrule
\end{tabular}
\end{table}

\newpage
\section{More Evaluation on Gaslighting Task}
\label{app:more_eval_on_galight}

To demonstrate the generalizability of SPAR, we extended our evaluation to three additional diverse benchmarks: MMMU, AI2Diagram, and MMBench. As presented in Table~\ref{tbl:result_on_multi_datasets_transposed}, this broader assessment confirms that the performance advantages of SPAR are consistent across varying task domains. On average, SPAR achieves a post-attack accuracy of 36.93\%, representing a significant improvement over both the LLaVA-v1.5 baseline (26.14\%) and GasEraser (33.34\%). This performance margin is maintained across every individual dataset, affirming that SPAR provides robust protection against induced hallucinations regardless of the underlying data distribution.

\begin{table}[ht!]
\centering
\caption{Transposed Evaluation on Induced Hallucination. ``Before" and ``After" denote model accuracy before and after gaslighting statements, respectively.}
\label{tbl:result_on_multi_datasets_transposed}
\begin{tabular}{clccc}
\toprule
Dataset & Metric & LLaVA-v1.5 & + GasEraser & + SPAR (Ours) \\
\midrule
\multirow{2}{*}{MMMU} 
& Before & 37.56 & 33.87 & 36.47 \\
& After  & 22.14 & 25.86 & \textbf{27.96} \\
\midrule
\multirow{2}{*}{AI2Diagram} 
& Before & 49.66 & 42.79 & 49.08 \\
& After  & 29.48 & 32.39 & \textbf{36.77} \\
\midrule
\multirow{2}{*}{MMBench} 
& Before & 72.07 & 68.02 & 72.19 \\
& After  & 26.80 & 41.77 & \textbf{46.05} \\
\midrule
\multirow{2}{*}{\textbf{Average}} 
& Before & 53.09 & 48.23 & 52.58 \\
& After  & 26.14 & 33.34 & \textbf{36.93} \\
\bottomrule
\end{tabular}
\end{table}

\newpage

\section{Detailed Results on POPE} 
\label{app:retails_on_pope}

Tables~\ref{tab:ms-coco}, \ref{tab:a-okvqa}, and \ref{tab:gqa} present the detailed evaluation results on the MS-COCO, A-OKVQA, and GQA subsets of the POPE benchmark, respectively.

\begin{table}[htbp]
\centering
\caption{Experimental results on MS-COCO dataset.}
\label{tab:ms-coco}
\small
\begin{tabular}{clcccc}
\toprule
\textbf{Split} & \textbf{Method} & \textbf{Acc} ($\uparrow$) & \textbf{Prec} ($\uparrow$) & \textbf{Rec} ($\uparrow$) & \textbf{F1} ($\uparrow$) \\
\midrule
\multirow{10}{*}{Random} & Regular & 83.13 & 81.94 & 85.00 & 83.44 \\
 & VCD & 87.00 & 86.13 & 88.20 & 87.15 \\
 & M3ID & 87.50 & 87.38 & 87.67 & 87.52 \\
 & AGLA & 89.20 & 88.89 & 89.60 & 89.24 \\
 & ONLY & 89.70 & 89.95 & 88.27 & 89.10 \\
 & VAR & 88.13 & 95.18 & 80.30 & 87.12 \\
 & Ours & 87.80 & \textbf{95.87} & 79.00 & 86.62 \\
 & Ours+AGLA & \textbf{89.83} & 89.23 & \textbf{90.60} & \textbf{89.91} \\
 & Ours+ONLY & 89.60 & 89.97 & 89.13 & 89.55 \\
 & Ours+VAR & 88.93 & 94.78 & 82.40 & 88.16 \\
\midrule
\multirow{10}{*}{Popular} & Regular & 81.17 & 78.28 & 86.27 & 82.08 \\
 & VCD & 83.10 & 79.96 & 88.33 & 83.94 \\
 & M3ID & 84.30 & 81.58 & 88.60 & 84.95 \\
 & AGLA & 85.50 & 82.93 & 89.40 & 86.04 \\
 & ONLY & 86.00 & 84.44 & 88.27 & 86.31 \\
 & VAR & 86.00 & 90.50 & 80.40 & 85.16 \\
 & Ours & 85.77 & \textbf{91.36} & 79.00 & 84.73 \\
 & Ours+AGLA & 85.63 & 82.53 & \textbf{90.40} & 86.29 \\
 & Ours+ONLY & \textbf{86.20} & 84.19 & 89.13 & \textbf{86.59} \\
 & Ours+VAR & 85.86 & 88.54 & 82.40 & 85.36 \\
\midrule
\multirow{10}{*}{Adversarial} & Regular & 77.43 & 73.31 & 86.27 & 79.26 \\
 & VCD & 77.17 & 72.18 & 88.40 & 79.47 \\
 & M3ID & 78.23 & 73.51 & 88.27 & 80.22 \\
 & AGLA & 79.37 & 74.19 & 90.07 & 81.36 \\
 & ONLY & 79.40 & 75.00 & 88.20 & 81.07 \\
 & VAR & 82.90 & 84.78 & 80.20 & 82.43 \\
 & Ours & \textbf{83.30} & \textbf{86.60} & 78.87 & \textbf{82.55} \\
 & Ours+AGLA & 79.23 & 73.97 & \textbf{90.20} & 81.29 \\
 & Ours+ONLY & 79.70 & 74.99 & 89.13 & 81.45 \\
 & Ours+VAR & 82.67 & 82.40 & 82.06 & 82.23 \\
\bottomrule
\end{tabular}
\end{table}

\begin{table}[h]
\centering
\caption{Experimental results on A-OKVQA dataset.}
\label{tab:a-okvqa}
\small
\begin{tabular}{clcccc}
\toprule
\textbf{Split} & \textbf{Method} & \textbf{Acc} ($\uparrow$) & \textbf{Prec} ($\uparrow$) & \textbf{Rec} ($\uparrow$) & \textbf{F1} ($\uparrow$) \\
\midrule
\multirow{10}{*}{Random} & Regular & 81.90 & 76.63 & 91.80 & 83.53 \\
 & VCD & 83.83 & 78.05 & 94.13 & 85.34 \\
 & M3ID & 84.67 & 79.25 & 93.93 & 85.97 \\
 & AGLA & 86.30 & 80.47 & \textbf{95.87} & 87.50 \\
 & ONLY & 86.07 & 80.91 & 94.40 & 87.14 \\
 & VAR & 89.27 & 90.34 & 87.93 & 89.12 \\
 & Ours & 89.23 & \textbf{90.73} & 87.40 & 89.03 \\
 & Ours+AGLA & 85.40 & 79.40 & 95.60 & 86.75 \\
 & Ours+ONLY & 86.87 & 81.78 & 94.87 & 87.84 \\
 & Ours+VAR & \textbf{89.37} & 89.34 & 89.40 & \textbf{89.37} \\
\midrule
\multirow{10}{*}{Popular} & Regular & 75.07 & 68.58 & 92.53 & 78.77 \\
 & VCD & 76.63 & 69.59 & 94.60 & 80.19 \\
 & M3ID & 77.80 & 70.98 & 94.07 & 80.91 \\
 & AGLA & 79.17 & 72.20 & 94.87 & 81.99 \\
 & ONLY & 79.00 & 72.17 & 94.40 & 81.80 \\
 & VAR & 84.20 & 81.67 & 88.20 & 84.81 \\
 & Ours & 84.43 & \textbf{82.50} & 87.40 & 84.88 \\
 & Ours+AGLA & 79.00 & 71.66 & \textbf{95.93} & 82.04 \\
 & Ours+ONLY & 79.17 & 72.20 & 94.87 & 81.99 \\
 & Ours+VAR & \textbf{84.70} & 81.91 & 89.10 & \textbf{85.30} \\
\midrule
\multirow{10}{*}{Adversarial} & Regular & 67.23 & 61.56 & 91.80 & 73.70 \\
 & VCD & 67.40 & 61.39 & 93.80 & 74.21 \\
 & M3ID & 68.60 & 62.22 & 94.73 & 75.11 \\
 & AGLA & 68.53 & 62.03 & 95.53 & 75.22 \\
 & ONLY & 68.70 & 62.35 & 94.40 & 75.70 \\
 & VAR & \textbf{77.73} & \textbf{72.93} & 88.20 & \textbf{79.84} \\
 & Ours & 77.27 & 72.67 & 87.40 & 79.36 \\
 & Ours+AGLA & 67.97 & 61.56 & \textbf{95.67} & 74.92 \\
 & Ours+ONLY & 69.07 & 62.62 & 94.60 & 75.36 \\
 & Ours+VAR & 76.10 & 70.73 & 89.10 & 78.84 \\
\bottomrule
\end{tabular}
\end{table}

\begin{table}[h]
\centering
\caption{Experimental results on GQA dataset.}
\label{tab:gqa}
\small
\begin{tabular}{clcccc}
\toprule
\textbf{Split} & \textbf{Method} & \textbf{Acc} ($\uparrow$) & \textbf{Prec} ($\uparrow$) & \textbf{Rec} ($\uparrow$) & \textbf{F1} ($\uparrow$) \\
\midrule
\multirow{10}{*}{Random} & Regular & 82.23 & 76.32 & 93.47 & 84.03 \\
 & VCD & 83.23 & 76.73 & 95.40 & 85.05 \\
 & M3ID & 84.20 & 78.00 & 95.27 & 85.77 \\
 & AGLA & 86.00 & 79.67 & 96.67 & 87.35 \\
 & ONLY & 86.70 & 80.94 & 96.00 & 87.83 \\
 & VAR & \textbf{89.00} & 89.16 & 88.80 & \textbf{88.98} \\
 & Ours & 88.90 & \textbf{89.78} & 87.80 & 88.78 \\
 & Ours+AGLA & 86.50 & 80.30 & \textbf{96.73} & 87.75 \\
 & Ours+ONLY & 86.47 & 80.73 & 95.80 & 87.62 \\
 & Ours+VAR & 88.70 & 88.93 & 88.40 & 88.67 \\
\midrule
\multirow{10}{*}{Popular} & Regular & 73.47 & 66.83 & 93.20 & 77.84 \\
 & VCD & 72.37 & 65.27 & 95.60 & 77.58 \\
 & M3ID & 73.87 & 66.70 & 95.33 & 78.49 \\
 & AGLA & 73.83 & 66.41 & 96.47 & 78.66 \\
 & ONLY & 74.03 & 66.70 & 96.00 & 78.71 \\
 & VAR & 81.93 & \textbf{78.18} & 88.60 & \textbf{83.06} \\
 & Ours & 81.63 & 78.16 & 87.80 & 82.70 \\
 & Ours+AGLA & 74.87 & 67.22 & \textbf{97.07} & 79.43 \\
 & Ours+ONLY & 74.93 & 67.59 & 95.80 & 79.26 \\
 & Ours+VAR & \textbf{82.87} & 77.89 & 88.53 & 82.87 \\
\midrule
\multirow{10}{*}{Adversarial} & Regular & 68.60 & 62.43 & 93.40 & 74.84 \\
 & VCD & 68.83 & 62.26 & 95.67 & 75.43 \\
 & M3ID & 68.67 & 62.16 & 95.40 & 75.28 \\
 & AGLA & 68.53 & 61.82 & \textbf{96.93} & 75.49 \\
 & ONLY & 69.23 & 62.55 & 95.87 & 75.70 \\
 & VAR & 79.13 & 74.50 & 88.60 & 80.94 \\
 & Ours & \textbf{79.50} & \textbf{75.30} & 87.80 & \textbf{81.10} \\
 & Ours+AGLA & 68.20 & 61.58 & 96.80 & 75.27 \\
 & Ours+ONLY & 69.47 & 62.70 & 96.13 & 75.89 \\
 & Ours+VAR & 78.50 & 73.50 & 89.00 & 80.52 \\
\bottomrule
\end{tabular}
\end{table}

\clearpage

\section{Qualitative Results} \label{appen:qualitive_results}
\subsection{Qualitative Results on GaslightingBench}
\begin{figure*}[h]
\centering
\includegraphics[width=0.7\linewidth]{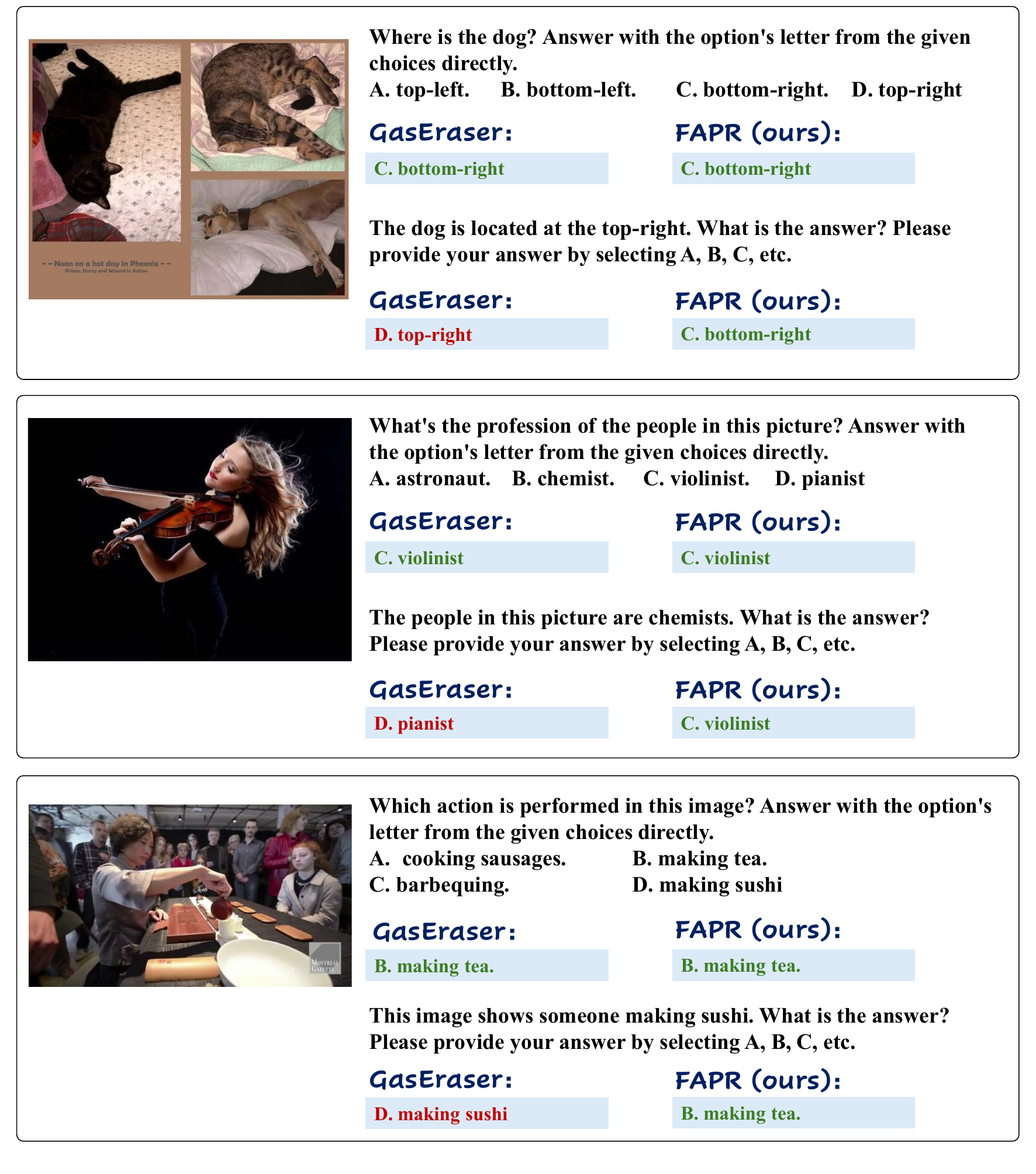}
\caption{Qualitative results on the GaslightingBench.}
\label{fig:qual_result_gas}
\end{figure*}

We present several qualitative results on GaslightingBench in Figure~\ref{fig:qual_result_gas}. The provided qualitative results demonstrate a comparative analysis of two models, \textbf{GasEraser} and \textbf{SPAR (ours)}, on visual question answering tasks, particularly when faced with misleading textual prompts. Across three distinct scenarios involving object location, profession identification, and action recognition, both models perform correctly on straightforward questions. However, a significant performance disparity emerges when the questions include a false premise. In every case with a misleading prompt---such as incorrectly stating a dog's location, a person's profession, or the action being performed---the \textbf{GasEraser} model is deceived and outputs an incorrect answer that aligns with the false textual information. In stark contrast, the \textbf{SPAR} model consistently ignores the misleading text and correctly answers based on the visual evidence in the image. This indicates that \textbf{SPAR} possesses a superior robustness and a stronger grounding in visual data, making it resilient to contradictory or adversarial textual prompts.

\end{document}